\documentclass{article} 
\usepackage{iclr2027_conference,times}

\usepackage{amsmath,amsfonts,bm}

\def\eqref#1{equation~\ref{#1}}

\def\1{\bm{1}}

\DeclareMathAlphabet{\mathsfit}{\encodingdefault}{\sfdefault}{m}{sl}
\SetMathAlphabet{\mathsfit}{bold}{\encodingdefault}{\sfdefault}{bx}{n}

\usepackage{url}
\usepackage{booktabs}
\usepackage{graphicx}
\usepackage{tikz-cd}
\usepackage{enumitem}
\usepackage{xcolor}         
\usepackage{makecell}
\usepackage{wrapfig}
\usepackage{mathtools}

\definecolor{iclrblue}{rgb}{0.21,0.49,0.74}
\usepackage[pagebackref,breaklinks,colorlinks,citecolor=iclrblue]{hyperref}

\title{Algebraic Interaction: A Recipe for Neural Architecture Generation}

\author{Haonan Dong
\\
Qiuzhen College, Tsinghua University\\
\texttt{dhn23@mails.tsinghua.edu.cn} \\
\And
Angelica I Aviles-Rivero~\thanks{Corresponding author} \\
YMSC, Tsinghua University \\
\texttt{aviles-rivero@tsinghua.edu.cn} \\
}

\iclrfinalcopy 
\begin{document}

\maketitle

\begin{abstract}

We introduce an algebraic foundation for neural architecture design in which convolution, attention, state-space models, and equivariant layers are expressed through compositions of bilinear product operators over two structures: a \emph{position algebra} governing positional routing and a \emph{feature algebra} governing feature interactions. This decomposition makes routing, feature algebra, and operator composition explicit design variables.
We focus on the feature-algebra axis through an \emph{Algebraic Recipe}: fix the routing and operator composition of an existing architecture while replacing its feature algebra. This preserves the information-flow structure while allowing computational and symmetry properties to be modified through algebraic specification.
We instantiate the recipe in linear attention, Mamba, and equivariant networks. Structure-constant and Clifford algebras improve accuracy at comparable parameter count in linear attention; structure-constant algebras yield up to $3\times$ faster training and inference in Vision Mamba at competitive accuracy; and an $\mathrm{SO}(3)$-representation algebra provides rotational equivariance by construction. These results demonstrate that feature algebra alone can control accuracy, efficiency, and symmetry without redesigning positional routing. 
This turns algebraic structure into an explicit design variable and reduces architecture design, within our framework, to algebraic specification.

\end{abstract}

\section{Introduction}
Deep-learning architectures are usually described by their high-level
mechanisms: convolution, attention, state-space models, and equivariant
layers. Each has its own computational structure and inductive bias. CNNs
\citep{lecun1998gradient} exploit locality and translation symmetry;
equivariant networks \citep{cohen2016group, thomas2018tensor} extend this
to general group symmetries; linear attention
\citep{katharopoulos2020transformersrnnsfastautoregressive} reduces the
quadratic cost of attention; and Mamba \citep{gu2023mamba} builds on
selective state-space models for efficient sequence modeling. These
mechanisms are typically taken as the building blocks of architecture
design. Despite their different forms, they share a common algebraic
structure.

We show that these high-level mechanisms are not primitives. They can be
understood through two elementary ingredients: (i) \emph{positional
routing}, which determines where information flows across positions; and
(ii) \emph{feature interaction}, which determines how transmitted
information is combined. We formalize these ingredients using a
\emph{position algebra}, which describes routing patterns, and a
\emph{feature algebra}, which describes feature interactions. Signals are
represented in the tensor product of these algebras, and high-level
mechanisms are generated through compositions of elementary bilinear
product operators that jointly implement routing and interaction.
This decomposition makes routing, feature interaction, and operator
composition explicit design variables. Rather than treating an
architecture as a monolithic mechanism, these components can be specified
separately and then integrated through operator composition. In this
paper, we focus on the feature-algebra axis. We fix the positional
routing and operator composition of an existing architecture while
replacing its feature algebra. This preserves the information-flow
structure while changing the algebraic structure of feature
interactions, allowing computational and symmetry properties to be
modified without redesigning positional routing.
Architecture design then becomes a problem of algebraic specification:
choose a routing pattern, choose a feature algebra encoding the desired
computational or symmetry properties, and compose the corresponding
operators. The resulting architecture inherits these algebraic properties
by construction---it is derived rather than discovered.

We formalize this construction as an \emph{Algebraic Recipe} and
instantiate it in linear attention, Mamba, and equivariant networks.
Structure-constant and Clifford algebras improve accuracy at comparable
parameter count in linear attention. In Vision Mamba, structure-constant
algebras yield up to $3\times$ faster training and inference while
retaining competitive accuracy. Replacing matrix algebra with an
$\mathrm{SO}(3)$-representation algebra yields rotational equivariance by
construction. These results show that changing the feature algebra alone
can alter accuracy--efficiency tradeoffs and impose symmetry without
redesigning positional routing, supporting our central hypothesis that
architectural mechanisms arise from the integration of routing and
feature interaction rather than constituting primitives themselves. \textbf{Our contributions are as follows.}

\begin{itemize}[leftmargin=*, nosep]
\item We introduce a tensor-product formalism
$\mathcal{B} \otimes \mathcal{A}$, where signals are elements of the
tensor product of a \emph{position algebra} $\mathcal{B}$ encoding
routing and a \emph{feature algebra} $\mathcal{A}$ encoding feature
interactions. We show that convolution, linear attention, Mamba, and
equivariant layers arise as compositions of elementary bilinear product
operators in this formalism. This identifies routing, feature algebra,
and operator composition as separate design degrees of freedom:
high-level mechanisms arise from their algebraic composition rather than
constituting primitives themselves.

\item We derive an \emph{Algebraic Recipe} for architecture construction:
keep the position algebra $\mathcal{B}$ and operator composition fixed,
replace the feature algebra $\mathcal{A}$, and derive the resulting
architecture. This separates positional routing from the algebra of
feature interaction, so that the feature algebra can be varied while
routing is held fixed. Properties encoded by the new feature algebra,
such as symmetry, are then inherited by the resulting feature
interactions by construction.

\item We instantiate the recipe with structure-constant algebras
$\mathcal{A}^{\lambda}$, Clifford algebras $\mathcal{A}^{Cl_N}$, and
$\mathrm{SO}(3)$-representation algebras
$\mathcal{A}^{\mathrm{SO}(3)}$. These constructions improve accuracy at
comparable parameter count in linear attention, yield up to $3\times$
faster training and inference in Vision Mamba at competitive accuracy,
and produce rotational equivariance by construction. Since positional
routing is unchanged in each case, these results isolate feature algebra
as a design variable that can alter accuracy--efficiency tradeoffs and
impose symmetry independently of routing.
\end{itemize}

\begin{figure}[t!]
  \centering
  \includegraphics[width=\linewidth]{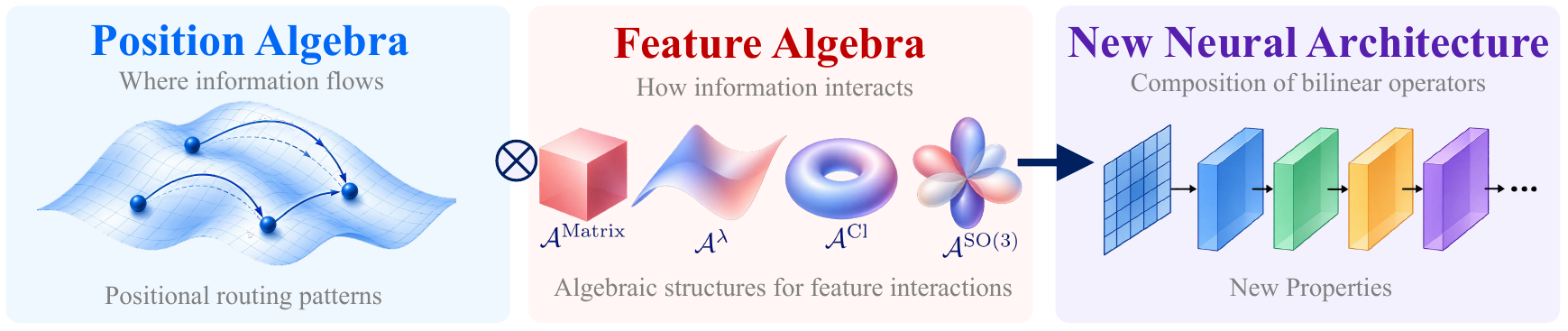}
  \caption{\textbf{Algebraic recipe for neural architecture generation.}
Position algebra determines where information flows, while feature algebra determines how information interacts. Their composition generates neural architectures with different computational and symmetry properties.}
  \label{fig:iclr_recipe}
\end{figure}

\section{Algebraic Formalism}
We introduce the algebraic formalism used throughout the paper. The
central observation is that the mechanisms underlying deep-learning
layers can be decomposed into two elementary ingredients: positional
routing and feature interaction. We model these using a position algebra
and a feature algebra, and represent signals in their tensor product. In
this formalism, high-level mechanisms are not primitives; they arise as
compositions of elementary product operators.

An algebra over $\mathbb{F}$, where
$\mathbb{F} \in \{\mathbb{R},\mathbb{C}\}$, is a vector space
$\mathcal{A}$ equipped with an $\mathbb{F}$-bilinear product
$
\mathcal{A}\times\mathcal{A}\to\mathcal{A},
 (a,b)\mapsto ab.
$
Positions---sequence indices, pixels, or spatial samples---are represented
in a vector space $\mathcal{B}$ with basis $\{f_a\}$, where
$f_a=f_{\vec r_a}$ corresponds to a sampled position $\vec r_a$.
Routing across positions is induced by the product law
$
f_a f_b
=
\sum_c
\lambda(\vec r_a,\vec r_b;\vec r_c)f_c.
$
We consider three routing mechanisms: convolution, local--local routing,
and local--global routing. Each corresponds to a particular choice of
$\lambda$ and is developed in Section~\ref{sec:position}.
Feature information is represented in a feature algebra $\mathcal{A}$,
whose product law determines feature--feature interactions. We consider
four choices: matrix algebra, structure-constant algebra, Clifford
algebra, and $\mathrm{SO}(3)$-representation algebra. These are developed
in Section~\ref{sec:feature}.
A signal is represented as an element of
$\mathcal{B}\otimes\mathcal{A}$, equipped with the product induced from
the two factors:
$
(f_a\otimes e_i)(f_b\otimes e_j)
=
(f_a f_b)\otimes(e_i e_j).
$
Thus a single multiplication combines positional routing through
$\mathcal{B}$ with feature interaction through $\mathcal{A}$, while the
tensor-product structure keeps the two components separate. See
Appendix~\ref{app: Tensor product} for the tensor-product construction.

The product in $\mathcal{B}\otimes\mathcal{A}$ defines the basic product
operator
$
\mathcal{O}^{\mathcal{B}\otimes\mathcal{A}}(X,Y)=XY,
$
where $XY$ denotes multiplication in
$\mathcal{B}\otimes\mathcal{A}$. Each product operator therefore couples
a routing mechanism, induced by $\mathcal{B}$, with a feature
interaction, induced by $\mathcal{A}$.
More complex mechanisms are obtained by composing these operators.
Given, for example,
$\mathcal{O}^{\mathcal{B}_1\otimes\mathcal{A}_1}$ and
$\mathcal{O}^{\mathcal{B}_2\otimes\mathcal{A}_2}$, their composition
successively applies the positional routing encoded by
$\mathcal{B}_1,\mathcal{B}_2$ and the feature interactions encoded by
$\mathcal{A}_1,\mathcal{A}_2$. More generally, a composition takes the
form
$
\Pi
=
\mathcal{O}^{\mathcal{B}_L\otimes\mathcal{A}_L}
\circ\cdots\circ
\mathcal{O}^{\mathcal{B}_1\otimes\mathcal{A}_1}.
$
Our central principle is that the \emph{mechanism of an architecture
emerges from positional routing and algebraic interaction of features}.
Complex mechanisms are therefore not introduced as primitive operations.
The product operator
$\mathcal{O}^{\mathcal{B}\otimes\mathcal{A}}$ is the elementary
interaction unit, and high-level mechanisms arise through compositions
of such units.
This decomposition separates the choice of positional routing, feature
algebra, and operator composition. In this paper, we focus on varying the
feature algebra while holding the positional routing and operator
composition fixed. This construction gives the \emph{Algebraic Recipe}.

\section{Position Algebra}\label{sec:position}
We now specify the positional component of the formalism. Recall that a
position algebra $\mathcal{B}$ is determined by a product
$
    f_a f_b
    =
    \sum_c \lambda(\vec r_a,\vec r_b;\vec r_c)f_c,
$
where the coefficients $\lambda$ determine how information is routed
between positions. Different choices of $\lambda$ recover different
information-flow patterns. We consider three cases used throughout the
paper: convolutional, local--local, and local--global routing. These
routing laws  later be held fixed while the feature algebra is
varied in the Algebraic Recipe.

\textbf{Convolution}
\label{sec: Convolution}
We first show how convolution is induced from product law $f_af_b = \sum_c\lambda(\vec{r}_a,\vec{r}_b;\vec{r_c})f_c$. The key of convolution is translation equivariance, that is for 
 translation operator   $\mathcal{T}_\Delta(f_{\vec{r}_a})=f_{\vec{r}_a+\Delta}$, and two elements $K,X\in \mathcal{B}\otimes\mathcal{A}$ with $K$ serving as kernel, we require $K(\mathcal{T}_{\Delta}(X))=\mathcal{T}_{\Delta}(KX)$. Then we get
 constraint  $\lambda(\vec{r}_a, \vec{r}_b; \vec{r}_c )=\lambda(\vec{r}_a, \vec{r}_b+\Delta; \vec{r}_c +\Delta )$.
After absorbing parameters, the only interesting solution is $\lambda(\vec{r}_a, \vec{r}_b; \vec{r}_c )=\delta_{\vec{r}_a, \vec{r}_b-\vec{r}_c}$. See \ref{app: convolution} for details.

Then for feature algebra $\mathcal{A}$ and two elements $K=\sum_a K(\vec{r}_a)f_a\in \mathcal{B}\otimes\mathcal{A}$, $X=\sum_b X(\vec{r}_b)f_b\in \mathcal{B}\otimes\mathcal{A}$ where $K(\vec{r}_a), \, X(\vec{r}_b) \in \mathcal{A}$, 
we have convolution  operator
\begin{equation*}
\mathcal{O}^{\mathcal{B}\otimes\mathcal{A}}(K,X)=KX=\sum_c\bigg[\sum_bK(\vec{r}_b-\vec{r}_c)X(\vec{r}_b)\bigg]f_c
\end{equation*}
The positional product produces convolutional routing, while
$K(\vec r_b-\vec r_c)X(\vec r_b)$ is multiplication in the feature
algebra $\mathcal{A}$. We denote this operator by
$\mathcal{O}^{\mathrm{Conv},\mathcal{A}}$.

\textbf{Local-Local Routing and Local-Global Routing}
The local-local routing is induced from product law $f_af_b = \delta_{\vec{r}_a,\vec{r}_b}f_a$. Then for two elements $X=\sum_a X(\vec{r}_a)f_a\in \mathcal{B}\otimes\mathcal{A}$, $Y=\sum_b Y(\vec{r}_b)f_b\in \mathcal{B}\otimes\mathcal{A}$ where $X(\vec{r}_a), \, Y(\vec{r}_b) \in \mathcal{A}$, we get basic operator 
\begin{equation*}
    \mathcal{O}^{\mathcal{B}\otimes\mathcal{A}}(X,Y)=XY=\sum_a\bigg[X(\vec{r}_a)Y(\vec{r}_a)\bigg]f_a
\end{equation*}
where $X(\vec{r}_a)Y(\vec{r}_a)$ is product in feature algebra $\mathcal{A}$. In this operator, local information of two components interact via algebraic interaction induced by feature algebra.

The local-global routing is induced from product law $f_af_b = f_a$. If there is causal condition in 1D sequence, the product law is  $f_af_b = 1_{b\leq a}f_a$.
For two elements $X=\sum_a X(\vec{r}_a)f_a\in \mathcal{B}\otimes\mathcal{A}$, $Y=\sum_b Y(\vec{r}_b)f_b\in \mathcal{B}\otimes\mathcal{A}$ where $X(\vec{r}_a), \, Y(\vec{r}_b) \in \mathcal{A}$, we get basic operator 
\begin{equation*}
    \mathcal{O}^{\mathcal{B}\otimes\mathcal{A}}(X,Y)=\sum_a\bigg[X(\vec{r}_a)\big(\sum_bY(\vec{r}_b)\big)\bigg]f_a \, \text{ or causal form } \sum_a\bigg[X(a)\big(\sum_{b\leq a}Y(b)\big)\bigg]f_a
\end{equation*}
In this operator, the local feature $X(\vec r_a)$ interacts with the
global aggregate $\sum_b Y(\vec r_b)$ through multiplication in the
feature algebra $\mathcal{A}$. We denote the local--local and
local--global operators by $\mathcal{O}^{L,\mathcal{A}}$ and
$\mathcal{O}^{G,\mathcal{A}}$, respectively.

These three position algebras provide the routing structures used in our
subsequent constructions; the feature algebra $\mathcal{A}$ remains
unspecified and can therefore be varied independently.

\section{Feature algebra} \label{sec:feature}
The position algebras of Section~\ref{sec:position} determine how
information is routed while leaving the feature algebra $\mathcal{A}$
unspecified. We now introduce the feature algebras used throughout the
paper and the interations induced by their product laws.


\textbf{$\mathcal{A}^{\text{Matrix}}$: Matrix Algebra}
The most commonly used algebra in deep learning is matrix algebra. Matrix algebra $\mathcal{A}^{\text{Matrix}}$ is defined with basis $\{e_{ij}\}$ and product $e_{ij}e_{kl}=\delta_{jk}e_{il}$. This is just matrix multiplication with basis $e_{ij}$ being matrix with $(i,j)$ element equals to $1$ and others equal to zero. Note that for non-square  matrix $M\in\mathbb{R}^{n\times r}$ and $N\in\mathbb{R}^{r\times m}$, they can  also be embed in $\mathcal{A}^{\text{Matrix}}$ as 
$\tilde{M}=\sum_{i=1}^n\sum_{j=1}^rM_{ij}e_{ij}$ and $\tilde{N}=\sum_{k=1}^r\sum_{l=1}^mN_{kl}e_{kl}$, then $\tilde{M}\tilde{N}$ in $\mathcal{A}^{\text{Matrix}}$ is usual matrix product $MN$. This include two special case: (i) matrix-vector product $m=1$, (ii) tensor product of vector $r=1$.

Take feature algebra to be $\mathcal{A}^{\text{Matrix}}$,
we get explicit basic operators. For example for data represented as $X=\sum_{a,i}X_i(\vec{r}_a)f_a\otimes e_{i1}\in\mathcal{B}\otimes\mathcal{A}^{\text{Matrix}}$ and filter  represented as $K=\sum_{a,i,j}K_{ij}(\vec{r}_a)f_a\otimes e_{ij}\in\mathcal{B}\otimes\mathcal{A}^{\text{Matrix}}$
we get convolution operator that transforms input channels to output channels
\begin{equation*}
\mathcal{O}^{\text{Conv},\mathcal{A}^{\text{Matrix}}}(K,X)=\sum_{c,i}\bigg[\sum_b\sum_jK_{ij}(\vec{r}_b-\vec{r}_c)X_j(\vec{r}_b)\bigg]f_c\otimes e_{i1}
\end{equation*}
This is the common form of convolution operator used in CNN.

\textbf{$\mathcal{A}^{\lambda}$:Algebra defined by structure constant}
Algebra $\mathcal{A}^{\lambda}$ is defined with basis $\{e_{i}\}$ and product law $e_{i}e_{j}=\sum_{k}\lambda_{ij}^ke_{k}$ where $\lambda_{ij}^k$ is learnable parameter called structure constant. See \ref{app: structure constant} for more discussion on structure constant. 
Take feature algebra to be $\mathcal{A}^{\lambda}$,
we get explicit basic operators. For example for data represented as $X=\sum_{a,i}X_i(\vec{r}_a)f_a\otimes e_{i}\in\mathcal{B}\otimes\mathcal{A}^{\lambda}$ and  $Y=\sum_{a,i}Y_{i}(\vec{r}_a)f_a\otimes e_{i}\in\mathcal{B}\otimes\mathcal{A}^{\lambda}$
we have local-local interaction operator
\begin{equation*}
\mathcal{O}^{L,\mathcal{A}^{\lambda}}(X,Y)=\sum_a\sum_k\bigg[\sum_{i,j}\lambda_{ij}^kX_i(\vec{r}_a)Y_{\textcolor{red}{j}}(\vec{r}_a)\bigg]f_a\otimes e_k
\end{equation*}
This gives a general bilinear feature interaction of the type used in
multiplicative-interaction models \citep{JayakumarCMSROT20}.

We can decompose $\lambda_{ij}^k=\sum_{r=1}^R U_{ir}^kV^k_{rj}$ for more efficient representation.
A special case is to take $R=1$ and $\lambda_{ij}^k=\delta_{ik}\delta_{jk}$. 
Then the product is $e_ie_j=\delta_{ij}e_i$. This is the usual Hadamard product used in deep learning and we write this algebra as $\mathcal{A}^{\text{Hadamard}}$.
Take feature algebra to be $\mathcal{A}^{\text{Hadamard}}$,
we get several explicit basic operators. For example for data represented as $X=\sum_{a,i}X_i(\vec{r}_a)f_a\otimes e_i\in\mathcal{B}\otimes\mathcal{A}^{\text{Hadamard}}$ and  $Y=\sum_{a,i}Y_i(\vec{r}_a)f_a\otimes e_i\in\mathcal{B}\otimes\mathcal{A}^{\text{Hadamard}}$,
we have local-local interaction operator
\begin{equation*}
\mathcal{O}^{L,\mathcal{A}^{\text{Hadamard}}}(X,Y)=\sum_a\sum_i\bigg[X_i(\vec{r}_a)Y_i(\vec{r}_a)\bigg]f_a\otimes e_i
\end{equation*}
If we nonlinear activate $Y_i(\vec{r}_a)$ with sigmoid function then $X_i(\vec{r}_a)\sigma(Y_i(\vec{r}_a))$ is gating and we write this gating operator as $\mathcal{O}_{\sigma}^{L,\mathcal{A}^{\text{Hadamard}}}$

Another example is local-global interaction operator $\mathcal{O}^{G,\mathcal{A}^{\text{Hadamard}}}(X,Y)=\sum_a\bigg[X_i(\vec{r}_a)\big(\sum_bY_i(\vec{r}_b)\big)\bigg]f_a $.
%
If the global aggregate is transformed by a two-layer MLP
$ F=\sigma\circ W_2\circ\delta\circ W_1,$ the resulting interaction
$ X_i(\vec r_a) F\!\left(\sum_bY_i(\vec r_b)\right)$ recovers the feature-wise gating used in the SE block~\citep{hu2019squeezeandexcitationnetworks}.

\textbf{$\mathcal{A}^{Cl_N}$: Clifford Algebra}
For vector space $\mathcal{V}$ with orthogonal  basis $\{e_n\}_{n=1}^N$, Clifford algebra $\mathcal{A}^{Cl_N}$  is defined with basis $\{1; \,e_{i_1}e_{i_2}\cdots e_{i_k},\, 1\leq i_1<i_2<\cdots <i_k\leq N\}$ and relation 
$e_ie_j+e_je_i=2\delta_{ij}$.
For example $\mathcal{A}^{Cl_2}$ is generated with basis $\{1, e_1, e_2, e_1e_2\}$.

By including $\mathcal{A}^{Cl_N}$ in feature algebra,
we get explicit basic operators. For example for data and filter  represented as  $X, K\in\mathcal{B}\otimes\mathcal{A}^{\text{Matrix}}\otimes\mathcal{A}^{Cl_2}$, we have Clifford convolution operator $\mathcal{O}^{\text{Conv},\mathcal{A}^{\text{Matrix}}\otimes\mathcal{A}^{Cl_2}}(K,X)=KX$. 
This recovers the Clifford neural layer of~\citep{brandstetter2023cliffordneurallayerspde} for PDE modeling.
See Appendix~\ref{app: Clifford convolution} for the full expression.

\textbf{$\mathcal{A}^{SO(3)}$: Algebra with $SO(3)$ Representation}
\label{sec: so(3)}
Let vector space $V$ be direct sum of irreducible $SO(3)$ representations,
$
   V = \bigoplus_{l\in\mathbb{N}} V_l,
   \;
   V_l = \mathrm{span}\{e^{l}_{m}\}_{m=-l,\dots,l},
$
We denote the linear representation  operator as $\rho(R)$ for $R\in SO(3)$. Product compatible with representation ($(\rho(R)e^{l_1}_{m_1})(\rho(R)e^{l_2}_{m_2})
    = \rho(R)\big(e^{l_1}_{m_1} e^{l_2}_{m_2}\big)$
) on $V$ is generated by Clebsch–Gordan coefficients $e^{l_1}_{m_1} e^{l_2}_{m_2}
   = \sum_{l}\sum_{m=-l}^{l}
        C^{lm}_{l_1 m_1,\,l_2 m_2}\,e^{l}_{m}$.

Take feature algebra to be $\mathcal{A}^{SO(3)}$,
we get explicit basic operators. For example for data represented as $X=\sum_{a;\,l,m}X_m^l(\vec{r}_a)f_a\otimes e_{m}^l\in\mathcal{B}\otimes\mathcal{A}^{SO(3)}$ and filter $K=\sum_{a;\, l,m}K_{m}^l(\vec{r}_a)f_a\otimes e_{m}^l\in\mathcal{B}\otimes\mathcal{A}^{SO(3)}$,
we have convolution operator
$\mathcal{O}^{\text{Conv},\mathcal{A}^{SO(3)}}(K,X)=KX$.
There is geometry $SO(3)$ transformation $\mathcal{T}_R$  (see \ref{app: geometry transformation}) defined on algebra $\mathcal{B}\otimes\mathcal{A}^{SO(3)}$. 
We want convolution operator $\mathcal{O}^{\text{Conv},\mathcal{A}^{SO(3)}}$ to be equivariant with respect to $\mathcal{T}_R$ that is $\mathcal{T}_R\big(\mathcal{O}^{\text{Conv},\mathcal{A}^{SO(3)}}(K,X)\big)=\mathcal{O}^{\text{Conv},\mathcal{A}^{SO(3)}}\big(K,\mathcal{T}_R(X)\big),\quad R\in SO(3)$.

This equivariance condition constrain $K_{m}^l(\vec{r})$ to take the form 
$K^{\,l}_{m}(\vec r)
   = R^{l}(\|\vec r\|)\,Y^{l}_{m}(\hat r),$
which recovers $ \mathcal{O}^{\text{Conv},\mathcal{A}^{SO(3)}}$ to be tensor field network introduced in \citep{thomas2018tensor}. See \ref{app: SO(3) equivariance} for details.

\section{Composition of Operators}
In the previous sections, we introduced basic operators
$\mathcal{O}^{\mathcal{B}\otimes\mathcal{A}}(X,Y)$, where
$\mathcal{B}$ induces convolution, local--local routing, or
local--global routing, and $\mathcal{A}$ determines the feature
interaction. These basic operators are the building blocks for more
complicated architectures.
Suppose we have two basic operators 
$\mathcal{O}^{\mathcal{B}_1\otimes\mathcal{A}_1}$, $\mathcal{O}^{\mathcal{B}_2\otimes\mathcal{A}_2}$.
When the output of $\mathcal{O}^{\mathcal{B}_1\otimes\mathcal{A}_1}$ is passed into $\mathcal{O}^{\mathcal{B}_2\otimes\mathcal{A}_2}$, the feature algebra  change from $\mathcal{A}_1$ to $\mathcal{A}_2$. We therefore
introduce a linear transfer map $\mathcal{W}:\mathcal{A}_1\to \mathcal{A}_2$. When $\mathcal{A}_1$ and $\mathcal{A}_2$ have the same representation,
$\mathcal{W}$ can simply be the identity. We can then compose the
operators as
\begin{equation*}
O=\mathcal{O}^{\mathcal{B}_2\otimes\mathcal{A}_2}(X,Y),\quad X\,\text{or}\, Y= \mathrm{id}\otimes \mathcal{W}(\mathcal{O}^{\mathcal{B}_1\otimes\mathcal{A}_1}(Z,W))
\end{equation*}
where $\mathrm{id}:\mathcal{B}_1\to\mathcal{B}_2$ is identity map as they share same underlying vector space.
We use linear attention and Mamba as examples to show how this
composition works.

\textbf{Compositional Construction of Linear Attention}
For data $Q(\vec{r}_a)=(Q_i(\vec{r}_a))\in \mathbb{R}^{1\times d}$, $K(\vec{r}_b)=(K_i(\vec{r}_b))\in \mathbb{R}^{d\times 1}$, $V(\vec{r}_b)=(V_i(\vec{r}_b))\in \mathbb{R}^{1\times d}$, linear attention \citep{katharopoulos2020transformersrnnsfastautoregressive} takes the
form $ O(\vec{r}_a)=\sum_b \big[Q(\vec{r}_a)K(\vec{r}_b)\big]V(\vec{r}_b)=\sum_bQ(\vec{r}_a)\big[K(\vec{r}_b)V(\vec{r}_b)\big]$

For Compositional Construction, we first prepare basic operators. Take feature algebra to be matrix algebra $\mathcal{A}^{\text{Matrix}}$. Then for $X=\sum_{a,i}X_i(a)f_a\otimes e_{i1}\in\mathcal{B}\otimes\mathcal{A}^{\text{Matrix}}$, $Y=\sum_{a,i}Y_{i}(a)f_a\otimes e_{1i}\in\mathcal{B}\otimes\mathcal{A}^{\text{Matrix}}$ and $Z=\sum_{a;\,i,j}Z_{ij}(a)f_a\otimes e_{ij}\in\mathcal{B}\otimes\mathcal{A}^{\text{Matrix}}$
we have local-local interaction  operator and local-global interaction operator
\begin{align*}
    \mathcal{O}_{\, \mathbb{R}^d\times\mathbb{R}^d\to\mathbb{R}^{d\times d}}^{L,\mathcal{A}^{\text{Matrix}}}(X,Y)&=\sum_{c;\,i,j}\bigg[X_i(\vec{r}_c)Y_j(\vec{r}_c)\bigg]f_c\otimes e_{ij}\\
     \mathcal{O}_{\, \mathbb{R}^d\times\mathbb{R}^{d\times d}\to\mathbb{R}^{d}}^{G,\mathcal{A}^{\text{Matrix}}}(Y,Z)&=\sum_{c;\,j}\bigg[\sum_iY_i(\vec{r}_c)\bigg(\sum_aZ_{ij}(\vec{r}_a)\bigg)\bigg]f_c\otimes e_{1j}
\end{align*}

Then linear attention is constructed as composition
\begin{equation}\label{eq: comp of linear attention}
    H= \mathcal{O}_{\, \mathbb{R}^d\times\mathbb{R}^d\to\mathbb{R}^{d\times d}}^{L,\mathcal{A}^{\text{Matrix}}}(K,V),\quad O= \mathcal{O}_{\, \mathbb{R}^d\times\mathbb{R}^{d\times d}\to\mathbb{R}^{d}}^{G,\mathcal{A}^{\text{Matrix}}}(Q,H)
\end{equation}

The compositional construction of linear attention shows that the key design idea is to control the position information flow together with suitable algebraic interaction of feature. The information flow is to first let local information of $K$ and $V$ flow into middle state $H$ via tensor product and then let global information of $H$ and local information of $Q$ flow into output via matrix product.

\textbf{Compositional Construction of Mamba}
Mamba \citep{gu2023mamba} is based on a state-space model.
We start from the linear ODE
\begin{equation*}
     \frac{dh_\alpha(t)}{dt}=A_\alpha(t)h_\alpha(t)+I_\alpha(t), \quad h_\alpha(t),I_\alpha(t)\in\mathbb{R}^{d_{\text{hidden}}},A_\alpha(t)\in \mathbb{R}^{d_{\text{hidden}}\times d_{\text{hidden}}}
\end{equation*}

The standard way to solve this linear ODE is multiply $e^{-\int_0^tA_\alpha(s)ds}$ then 
$h_\alpha(t)$ is solved by (suppose $h_\alpha(0)=0$ )
$h_\alpha(t)=e^{\int_0^{t}A_\alpha(s)ds}\int_0^t e^{-\int_0^{\tau}A_\alpha(s)ds}I_\alpha(\tau)d\tau$.
We consider $A_\alpha$ to be diagonal $A_\alpha=\text{diag}(A_{\alpha1},...,A_{\alpha d_{\text{hidden}}})$. 
Then in component of $h_\alpha=(h_{\alpha i})$,  $i = 1,...,d_{\text{hidden}}$, we get 
\begin{equation*}
h_{\alpha i}(t)=e^{\int_0^{t}A_{\alpha i} (s)ds}\int_0^t e^{-\int_0^{\tau}A_{\alpha i} (s)ds}I_{\alpha i} (\tau)d\tau \xrightarrow{\text{discrete}} e^{\sum_{l=1}^tA_{\alpha i} (l)}\sum_{k=1}^t e^{-\sum_{l=1}^kA_{\alpha i} (l)}I_{\alpha i} (k)
\end{equation*}

In Mamba, for input $X_\alpha(k)$, the quantities
$A_{\alpha i}(k)$ and $I_{\alpha i}(k)$ are parameterized as
\begin{equation*}
    \begin{cases}
    \Delta_{\alpha i}(k)=\Delta_{\alpha }(k)=\text{softplus}(\sum_\beta W^{\Delta}_{\alpha\beta}X_\beta(k)),\quad A_{\alpha i}(k) = \lambda_{\alpha i}\Delta_{\alpha i}(k),\\
    I_{\alpha i}(k)=\Delta_{\alpha i}(k)B_i(k)X_\alpha(k),\quad B_i(k)=\sum_\beta W^B_{i\beta}X_\beta(k)
    \end{cases}
\end{equation*}

The final output is:
$
    Y_\alpha(t)=\sum_ih_{\alpha i}(t)C_i(t),\quad C_i(t)=\sum_\beta W^C_{i \beta}X_\beta(t)
$

The compositional construction is the following. 
\begin{equation}\label{eq: comp of mamba}
    \begin{cases}
        I_0=\mathcal{O}_{\, \mathbb{R}^d\times\mathbb{R}^d\to\mathbb{R}^{d\times d}}^{L,\mathcal{A}^{\text{Matrix}}}(X,B),\quad  I=\mathcal{O}_{\, \mathbb{R}^{d\times d}\times\mathbb{R}^{d\times d}\to\mathbb{R}^{d\times d}}^{L,\mathcal{A}^{\text{Hadamard}}}(\Delta,I_0),\quad A=\mathcal{O}^{\mathcal{A}^\text{Hadamard}}_{\mathbb{R}^{d\times d}}(\Delta,\lambda),\\
H^{\text{local}}=\mathcal{O}^{G,\mathcal{A}^\text{Hadamard}}_{\exp(-),\mathbb{R}^{d\times d}\times\mathbb{R}^{d\times d}\to\mathbb{R}^{ d\times d}}(I,A), \quad H^{\text{global}}=\mathcal{O}^{G,\mathcal{A}^\text{Hadamard}}_{\mathbb{R}^{d\times d}}(1,H^{\text{local}}),  \\
H=\mathcal{O}^{G,\mathcal{A}^\text{Hadamard}}_{\exp,\mathbb{R}^{d\times d}\times\mathbb{R}^{d\times d}\to\mathbb{R}^{d\times d}}(H^{\text{global}},A),\quad Y = \mathcal{O}^{L,\mathcal{A}^{\text{Matrix}}}_{\mathbb{R}^{d\times d}\times\mathbb{R}^{d}\to\mathbb{R}^{d}}(H,C)
    \end{cases}
\end{equation}
See \ref{app: Mamba} for details. The key design idea behind this construction is also position routing combined with feature interaction. 
The first two local information flow construct $I$ via  tensor product and Hadamard product. The three global information flow to construct $H$ reflects three integration in solving SSM ODE. The final output is through matrix product interaction between $H$ and $C$.

\section{Architectures from the Algebraic Recipe}
\label{sec: architectures_recipe}

We now apply the Algebraic Recipe by keeping the positional routing and
operator composition fixed while changing the feature algebra. We first
replace matrix-valued interactions by structure-constant and Clifford
interactions in linear attention and Mamba. We then consider an
$\mathrm{SO}(3)$-representation algebra to incorporate rotational
symmetry into the feature interaction.

For two feature vectors $X$ and $Y$, the interaction in
$\mathcal{A}^{\lambda}$ is
$\sum_{ij}\lambda_{ij}^{k}X_iY_j$, whereas the corresponding interaction
in $\mathcal{A}^{\mathrm{Matrix}}$ is the matrix-valued product
$X_iY_j$. Thus $\mathcal{A}^{\lambda}$ maps the bilinear interaction
directly back into a vector feature space rather than explicitly
maintaining a $d\times d$ matrix-valued feature. We use this difference
in the following constructions.

\subsection{$\lambda$-Clifford Interaction}
\label{sec: lambda_clifford}

As shown in ~\eqref{eq: comp of linear attention}, linear attention
is constructed as the composition
$H=
\mathcal{O}_{\,\mathbb{R}^d\times\mathbb{R}^d
\to\mathbb{R}^{d\times d}}^{L,\mathcal{A}^{\mathrm{Matrix}}}(K,V),
\quad
O=
\mathcal{O}_{\,\mathbb{R}^d\times\mathbb{R}^{d\times d}
\to\mathbb{R}^{d}}^{G,\mathcal{A}^{\mathrm{Matrix}}}(Q,H)$.
We keep the positional information flow fixed while replacing the
feature algebras by $\mathcal{A}^{\lambda}$ and
$\mathcal{A}^{Cl_d}$. Specifically, we consider the basic operators
$\mathcal{O}_{\,
\mathbb{R}^d\times\mathbb{R}^d\to\mathbb{R}^{d}
    }^{L,\mathcal{A}^{\lambda}}(X,Y)
    =
    \sum_{c;\,k}
    \bigg[
    \sum_{i,j}\lambda_{ij}^k
    X_i(\vec r_c)Y_j(\vec r_c)
    \bigg]f_c\otimes e_k,$ and 
\begin{align*}
    \mathcal{O}_{\,
    \mathbb{R}^d\times\mathbb{R}^{d}\to\mathbb{R}^{d_2}
    }^{G,\mathcal{A}^{Cl_d}}(X,Y)
    &=
    \sum_c
    \bigg[
    \sum_i X_i(\vec r_c)
    \bigg(\sum_bY_i(\vec r_b)\bigg)
    \bigg]f_c\otimes e_0
    \\
    &\quad+
    \sum_c\sum_{i<j}
    \bigg[
    X_i(\vec r_c)
    \bigg(\sum_bY_j(\vec r_b)\bigg)
    -
    X_j(\vec r_c)
    \bigg(\sum_bY_i(\vec r_b)\bigg)
    \bigg]f_c\otimes e_i e_j.
\end{align*}


We then compose them in the same way as in linear attention:
\begin{equation}
\label{eq: lambda_clifford}
H=
\mathcal{O}_{\,
\mathbb{R}^d\times\mathbb{R}^d\to\mathbb{R}^{d}
}^{L,\mathcal{A}^{\lambda}}(K,V),
\quad
O=
\mathcal{O}_{\,
\mathbb{R}^d\times\mathbb{R}^{d}\to\mathbb{R}^{d_2}
}^{G,\mathcal{A}^{Cl_d}}(Q,H).
\end{equation}
Here
$
d_2=1+\frac{d(d-1)}{2},
$
and the output $O$ is projected from dimension $d_2$ back to dimension
$d$. For the structure constants, we use the factorization
$
\lambda_{ij}^k
=
\sum_{r=1}^{R}U_{ir}^kV_{rj}^k,
$
and investigate the effect of the bilinear rank $R$.

\textbf{Evaluation.}
To evaluate linear attention and the newly constructed
$\lambda$-Clifford Interaction, we consider a ResNet20 architecture
\citep{he2015deepresiduallearningimage} and replace convolution with
either a linear-attention block or a $\lambda$-Clifford Interaction
block, following the convolution-to-attention replacement used in SASA
\citep{ramachandran2019standaloneselfattentionvisionmodels}. For the
$\lambda$-Clifford model, all other components are kept fixed relative
to the linear-attention baseline, so that the main architectural change
is the feature interaction induced by the new feature algebras.
We train the models on CIFAR-10 and report mean test accuracy and
standard deviation over five random seeds. Full architectural details,
data preprocessing, optimization settings, and checkpoint-selection
protocol are given in
Appendix~\ref{app: Experiment lambda-Clifford}. The results are shown in
Table~\ref{tab: lamdba-Clifford}.
\begin{table}[htbp]
\centering
\caption{Comparison of $\lambda$-Clifford Interaction with ResNet20 and
linear attention on CIFAR-10. Replacing the matrix feature interaction
in linear attention with the proposed $\lambda$-Clifford Interaction
improves accuracy at a comparable parameter count.}
\label{tab: lamdba-Clifford}
\small
\resizebox{0.95\linewidth}{!}{
\begin{tabular}{lcccc}
\toprule
\textbf{Model}
& ResNet20
& Linear Attention
& $\lambda$-Clifford ($R=1$)
& $\lambda$-Clifford ($R=4$) \\
\midrule
Accuracy
& $89.670\pm0.250$
& $89.560\pm0.498$
& $90.356\pm0.273$
& $90.456\pm0.154$ \\
Parameters
& 272K
& 195K
& 201K
& 202K \\
\bottomrule
\end{tabular}
}
\end{table}

Both $\lambda$-Clifford models outperform the linear-attention baseline
at a comparable parameter count. The $R=1$ model achieves
$90.356\pm0.273$ accuracy with 201K parameters, compared with
$89.560\pm0.498$ for linear attention with 195K parameters. Increasing
the rank to $R=4$ gives $90.456\pm0.154$ accuracy with 202K parameters.
The $\lambda$-Clifford models also outperform the ResNet20 baseline in
this experiment.

These results support several observations about the role of feature
algebra. First, the matrix-valued interaction $X_iY_j$ of
$\mathcal{A}^{\mathrm{Matrix}}$ can be replaced by the bilinear
interaction
$
    \sum_{i,j}\lambda_{ij}^{k}X_iY_j
$
of $\mathcal{A}^{\lambda}$ without degrading performance in this
experiment. The corresponding intermediate representation changes from
a matrix-valued feature,
$
    \mathbb{R}^d\times\mathbb{R}^d
    \longrightarrow
    \mathbb{R}^{d\times d},
$
to a vector-valued feature,
$
    \mathbb{R}^d\times\mathbb{R}^d
    \longrightarrow
    \mathbb{R}^{d}.
$
The resulting feature can then be composed directly with a different
feature algebra, here the Clifford algebra
$\mathcal{A}^{Cl_d}$. Thus the construction changes not only the
dimension of the intermediate interaction but also the algebra through
which subsequent feature interactions are formed.
Second, the experiment separates this change in feature interaction from
the positional routing. The $\lambda$-Clifford construction preserves
the local--local and local--global routing pattern of the
linear-attention composition while replacing its matrix feature
algebra. Nevertheless, the resulting architecture remains competitive
and in this experiment achieves higher accuracy. This provides direct
empirical evidence that the routing structure can be retained while the
feature algebra is changed substantially.

More importantly, the resulting architecture no longer admits the
standard interpretation of linear attention in terms of attention
scores: the matrix-valued interaction has been replaced by
$\mathcal{A}^{\lambda}$ and $\mathcal{A}^{Cl_d}$ interactions.
Nevertheless, the architecture preserves the information-flow structure
of the original composition and improves its performance in this
experiment. This supports the central claim of our framework that the
high-level mechanism need not be treated as a primitive. The same
positional routing can be integrated with different algebraic feature
interactions to produce a different architecture with competitive
performance.

\subsection{$\lambda$-SSM}
\label{sec: lambda_ssm}

We next apply the same recipe to the compositional construction of Mamba
in ~\eqref{eq: comp of mamba}. We keep its positional routing and
Hadamard interactions fixed and replace the matrix interactions used to
construct the input and output by $\mathcal{A}^{\lambda}$. Explicitly,
$\mathcal{O}_{\mathbb{R}^d\times\mathbb{R}^d\to\mathbb{R}^{d}}^{L,\mathcal{A}^{\lambda}}(B,X)=\sum_{k,i}\left[\sum_{\alpha,\beta}\lambda_{\alpha\beta}^{i}B_\beta(k)X_\alpha(k)\right]f_k\otimes e_i,$ and $\mathcal{O}_{\mathbb{R}^d\times\mathbb{R}^d\to\mathbb{R}^{d}}^{L,\mathcal{A}^{\lambda}}(H,C)=\sum_{k,\alpha}\left[\sum_{i,\beta}\lambda_{i\beta}^{\alpha}H_i(k)C_\beta(k)\right]f_k\otimes e_\alpha.$
Keeping the remaining routing and Hadamard interactions unchanged gives
\begin{equation}\label{eq: lambda-SSM}
    \begin{cases}
        I_0=\mathcal{O}_{\, \mathbb{R}^d\times\mathbb{R}^d\to\mathbb{R}^{d}}^{L,\mathcal{A}^{\lambda}}(B,X),\quad  I=\mathcal{O}_{\, \mathbb{R}^{d}\times\mathbb{R}^{d}\to\mathbb{R}^{ d}}^{L,\mathcal{A}^{\text{Hadamard}}}(\Delta,I_0),\quad A=\mathcal{O}^{\mathcal{A}^\text{Hadamard}}_{\mathbb{R}^{d}}(\Delta,\lambda),\\
H^{\text{local}}=\mathcal{O}^{G,\mathcal{A}^\text{Hadamard}}_{\exp(-),\mathbb{R}^{d}\times\mathbb{R}^{d}\to\mathbb{R}^{ d}}(I,A), \quad H^{\text{global}}=\mathcal{O}^{G,\mathcal{A}^\text{Hadamard}}_{\mathbb{R}^{d}}(1,H^{\text{local}})\\ H=\mathcal{O}^{G,\mathcal{A}^\text{Hadamard}}_{\exp,\mathbb{R}^{d}\times\mathbb{R}^{d}\to\mathbb{R}^{d}}(H^{\text{global}},A), \quad
     Y=\mathcal{O}_{\, \mathbb{R}^d\times\mathbb{R}^d\to\mathbb{R}^{d}}^{L,\mathcal{A}^{\lambda}}(H,C)
    \end{cases}
\end{equation}

\textbf{Evaluation.}
We adopt a Vision Mamba architecture
\citep{zhu2024visionmambaefficientvisual} and replace its Mamba
dynamical layer with the constructed $\lambda$-SSM layer. All other
components of the Vision Mamba architecture are kept fixed, and the
$\lambda$-SSM model is trained using the same hyperparameters as the
Vision Mamba baseline. We evaluate the models on CIFAR-10. Training and
inference speeds are measured on the same hardware under identical
experimental configurations; full architectural and training details
are given in Appendix~\ref{app: experiment lambda SSM}.
\begin{table}[htbp]
\centering
\caption{Accuracy and computational speed of Vision Mamba and
$\lambda$-SSM variants on CIFAR-10. Replacing the matrix-valued
interaction with $\mathcal{A}^{\lambda}$ yields up to approximately
$3\times$ faster training and inference while retaining competitive
accuracy.}
\label{tab: lambdalambda}
\small
\begin{tabular}{lccc}
\toprule
\textbf{Model} & \textbf{Accuracy} & \textbf{Train (it/s)} & \textbf{Infer. (it/s)} \\
\midrule
Mamba
& $86.818\pm0.141$ & 4.81 & 15 \\
$\lambda$-SSM (increased $H$)
& $86.710\pm0.354$ & 13.96 & 43 \\
$\lambda$-SSM (full)
& $86.350\pm0.150$ & 14.02 & 45 \\
$\lambda$-SSM ($R=1$)
& $85.536\pm0.485$ & -- & -- \\
$\lambda$-SSM ($R=3$)
& $85.734\pm0.388$ & -- & -- \\
\bottomrule
\end{tabular}
\end{table}

The full $\lambda$-SSM achieves $86.350\pm0.150$ accuracy, compared
with $86.818\pm0.141$ for the original Mamba model. At the same time,
training speed increases from $4.81$ to $14.02$ iterations/s and
inference speed from $15$ to $45$ iterations/s, corresponding to
approximately $2.9\times$ and $3.0\times$ speedups, respectively.

The computational difference follows the change in feature interaction.
In the Mamba construction, the matrix product produces a matrix-valued
hidden state with feature dimension
$d_{\mathrm{head}}\times d_{\mathrm{hidden}}$. In $\lambda$-SSM, the
bilinear interaction
    $\sum_{i,j}\lambda_{ij}^{k}X_iY_j$
maps directly back into the feature space. In our implementation,
$d_{\mathrm{head}}=64$ and $d_{\mathrm{hidden}}=16$, so the
matrix-valued Mamba state has dimension $64\times16=1024$ per head,
whereas the $\lambda$-SSM hidden state has dimension $16$ per head.
Thus replacing the matrix interaction by $\mathcal{A}^{\lambda}$
substantially reduces the dimension of the intermediate interaction.
The measured speedups in Table~\ref{tab: lambdalambda} show the
computational effect of this replacement in the present implementation.

The vector-valued hidden state also allows additional feature
interactions to be composed directly with $H$. In particular, we
construct a self-interaction of $H$ using the Hadamard algebra, which
serves as a gating interaction: $\widetilde{H}
=
\mathcal{O}_{
\sigma,\mathbb{R}^d\times\mathbb{R}^d\to\mathbb{R}^d
}^{L,\mathcal{A}^{\mathrm{Hadamard}}}
\bigl(\mathcal{W}_1(H),\mathcal{W}_2(H)\bigr),
\quad
Y=
\mathcal{O}_{
\,\mathbb{R}^d\times\mathbb{R}^d\to\mathbb{R}^{d}
}^{L,\mathcal{A}^{\lambda}}(\widetilde{H},C).$
This increases the order of the algebraic interaction as in
\citep{dong2026productinteractionalgebraicformalism}. The
``increased $H$'' column of Table~\ref{tab: lambdalambda} reports this
construction. It reaches $86.710\pm0.354$ accuracy while retaining
nearly the same training and inference speed as the full
$\lambda$-SSM.
More importantly, this experiment tests the separation between
positional routing and feature interaction in the Algebraic Recipe.
The positional routing and Hadamard components inherited from the SSM
construction remain fixed, while the matrix feature interaction is
replaced by $\mathcal{A}^{\lambda}$. Despite the substantial change in
the hidden-state representation, the resulting architecture retains
competitive performance while changing the computational tradeoff.
Together with the $\lambda$-Clifford construction, this shows that the
same routing and composition structure can support different feature
algebras with different accuracy--efficiency tradeoffs.

\textbf{Effect of Bilinear Rank.}  The algebraic interaction in $\mathcal{A}^{\lambda}$ between feature
vectors $X$ and $Y$ is determined by the bilinear form
$
    \sum_{i,j}\lambda_{ij}^kX_iY_j.
$
For the factorization
$
    \lambda_{ij}^k
    =
    \sum_{r=1}^{R}U_{ir}^kV_{rj}^k,
$
the rank of the corresponding bilinear form is at most $R$. Increasing
$R$ therefore increases the capacity of this bilinear form to represent
different channel-interaction patterns.
Empirically, however, the importance of $R$ depends on the algebraic
interaction in which $\mathcal{A}^{\lambda}$ is used. For
$\lambda$-SSM, Table~\ref{tab: lambdalambda} shows a clear dependence
on rank. The full-rank model reaches $86.350\pm0.150$, compared with
$85.536\pm0.485$ for $R=1$ and $85.734\pm0.388$ for $R=3$.
Increasing $R$ from $1$ to $3$ gives a modest improvement, while the
full-rank interaction performs substantially better than either
low-rank variant.
For $\lambda$-Clifford Interaction, the dependence on $R$ is much
weaker. As shown in Table~\ref{tab: lamdba-Clifford}, $R=1$ already
reaches $90.356\pm0.273$, compared with $90.456\pm0.154$ for $R=4$.
This suggests that the required capacity of the
$\mathcal{A}^{\lambda}$ interaction depends on the feature algebra with
which it is composed. In the $\lambda$-Clifford construction, the
subsequent Clifford interaction may provide additional feature mixing,
reducing the sensitivity to the rank of $\lambda_{ij}^k$. A systematic
study of the interaction between bilinear rank and feature-algebra
composition is left for future work.

\subsection{Symmetry and Clebsch--Gordan Interaction}
\label{sec: Clebsch–Gordan Interaction}

Symmetry provides an important source of inductive bias in neural
architectures. In our formalism, symmetry can constrain either
positional routing or feature interaction. Let $\mathcal{T}_R$ denote a
symmetry transformation. An operator $\mathcal{O}$ is equivariant with
respect to this transformation if
$
    \mathcal{O}\circ\mathcal{T}_R
    =
    \mathcal{T}_R\circ\mathcal{O}.
$
As shown in Section~\ref{sec: Convolution}, translation equivariance
constrains the product law of the position algebra and leads to
convolutional routing. In Section~\ref{sec: so(3)}, rotational
equivariance constrains the interaction of $\mathrm{SO}(3)$
representations and gives the Clebsch--Gordan product used in Tensor
Field Networks.

We use this structure to test another consequence of the Algebraic
Recipe. We construct two architectures with the same positional routing
and the same operator-composition structure but different feature
algebras. Starting from the linear-attention composition in~\eqref{eq: comp of linear attention}, we replace
$\mathcal{A}^{\mathrm{Matrix}}$ by the
$\mathrm{SO}(3)$-representation algebra
$\mathcal{A}^{\mathrm{SO}(3)}$. The resulting feature interactions are
Clebsch--Gordan products:
\begin{equation}
\label{eq: cg_interaction_main}
H=
\mathcal{O}^{L,\mathcal{A}^{\mathrm{SO}(3)}}(K,V),
\qquad
O=
\mathcal{O}^{G,\mathcal{A}^{\mathrm{SO}(3)}}(Q,H).
\end{equation}
We retain the notation $Q,K,V$ only to make the correspondence with the
linear-attention composition explicit; the resulting operators are not
interpreted as an attention mechanism.
Here the positional routing and operator-composition structure are
unchanged, while the feature interaction respects the
$\mathrm{SO}(3)$ representation structure. The full
Clebsch--Gordan expansion is given in
Appendix~\ref{app:experiment_Clebsch–Gordan}.

\textbf{Evaluation.}
We evaluate the construction on the QM9 missing-point task. To isolate
the effect of the algebraic mixing rule, the Clebsch--Gordan and
matrix-interaction models use the same architecture outside the feature
interaction. 
All other components are kept the same.
Full architectural details, training curves, and additional results are
given in Appendix~\ref{app:experiment_Clebsch–Gordan}.

\begin{wraptable}{r}{0.48\textwidth}
\vspace{-10pt}
\centering
\caption{Accuracy by distance on the QM9 missing-point task for
Clebsch--Gordan and matrix interactions.}
\label{tab:accuracy_distance_small}
\small
\setlength{\tabcolsep}{4pt}
\begin{tabular}{lcc}
\toprule
\textbf{Atoms}
& \shortstack{\textbf{Clebsch--Gordan}\\\textbf{(epoch 10)}}
& \shortstack{\textbf{Matrix}\\\textbf{(epoch 100)}} \\
\midrule
5--18 & 0.9396 & 0.3493 \\
19    & 0.9567 & 0.3146 \\
23    & 0.9587 & 0.2836 \\
29    & 0.9532 & 0.2554 \\
\bottomrule
\end{tabular}
\vspace{-8pt}
\end{wraptable}
The Clebsch--Gordan interaction substantially outperforms the
matrix-interaction baseline across all four molecule-size groups in
this experiment, despite being evaluated after 10 epochs compared with
100 epochs for the matrix-interaction baseline. 
This comparison illustrates a different role of feature algebra from
the previous two constructions. In $\lambda$-Clifford Interaction and
$\lambda$-SSM, changing the feature algebra changes the intermediate
representation and the resulting accuracy--efficiency tradeoff. Here
the feature algebra encodes a symmetry constraint. Clebsch--Gordan
products preserve the $\mathrm{SO}(3)$ representation structure,
whereas flattening the representation and channel dimensions before
applying a matrix interaction does not preserve this structure in the
same way.

The experiment therefore connects the algebraic construction directly
to inductive bias. The positional routing and operator-composition
structure are held fixed, while changing the feature algebra changes
whether the feature interaction respects the underlying representation
structure. Together with the previous experiments, this supports the
separation of positional routing and feature algebra as distinct design
variables: routing determines where information is exchanged, while
the feature algebra determines how the transmitted features interact
and can encode structural constraints such as symmetry.

\section{Conclusion}
We introduced an algebraic framework in which neural architectures are
constructed from positional routing, feature interaction, and operator
composition. This decomposition shows that high-level mechanisms need
not be treated as primitives: fixing the routing and composition while
changing the feature algebra produces architectures with different
accuracy--efficiency tradeoffs and symmetry properties. Our
$\lambda$-Clifford, $\lambda$-SSM, and Clebsch--Gordan constructions
demonstrate this principle across attention, state-space, and
equivariant architectures. The main contribution is therefore not a
single new architecture, but a change in the level at which
architectures can be designed: algebraic structure becomes an explicit
design variable, and architecture construction becomes a problem of
algebraic specification.
\nopagebreak

\bibliography{iclr2027_conference}
\bibliographystyle{iclr2027_conference}

\appendix

\section{Related Work}

\paragraph{Algebraic perspectives on deep learning.}
Several works have attempted to understand deep learning from an algebraic perspective. \citep{gavranovic2024categorical} developed a categorical framework for describing the high-level structure of neural networks. While this provides a powerful abstract language, it is not intended as a practical framework for architecture construction. 
More recently, \citep{dong2026productinteractionalgebraicformalism} introduced an algebraic formalism for deriving neural architectures through product interaction operators. However, its notion of interaction order is relatively coarse and does not explicitly consider detailed interaction structure. In contrast, our framework explicitly introduces algebraic structure for positional routing and feature interaction, providing a structured compositional framework that naturally leads to architecture generation.

\paragraph{Multiplicative interactions.}
Multiplicative feature interactions have been widely studied in deep learning. \citep{JayakumarCMSROT20} proposed bilinear interactions as an expressive mechanism for modeling feature dependencies, while \citep{Chrysos_2021} further generalized this idea to higher-order polynomial interactions. Our work takes a different perspective. Instead of introducing interaction functions directly, we regard multiplicative interactions as arising naturally from the product law of a feature algebra. This algebraic viewpoint provides a principled way to construct higher-order interactions while jointly modeling feature interaction and positional routing through operator composition.

\paragraph{Clifford algebra.}
Clifford algebra has recently been employed as a powerful representation for geometric feature interactions \citep{brandstetter2023cliffordneurallayerspde,ji2026cliffordnetneedgeometricalgebra}. In our framework, Clifford algebra is not viewed as a standalone architecture, but as one possible choice of feature algebra. This interpretation places Clifford-based models within a unified algebraic framework together with other feature interaction mechanisms.

\paragraph{Equivariant learning.}
Encoding symmetry as an inductive bias has been a central theme in neural architecture design. Convolutional neural networks exploit translation symmetry, while group-equivariant networks generalize this idea to broader transformation groups \citep{cohen2016group}. Among these approaches, architectures based on irreducible group representations are most closely related to our work. Harmonic Networks \citep{worrall2017harmonicnetworksdeeptranslation} employ irreducible representations to preserve two-dimensional rotational symmetry, whereas Tensor Field Networks (TFNs) \citep{thomas2018tensor} and related models \citep{kondor2018clebsch} extend this framework to three-dimensional rotations using Clebsch--Gordan coefficients to decompose tensor-product representations into irreducible representations.

Our interpretation differs fundamentally from this representation-theoretic viewpoint. Rather than viewing Clebsch--Gordan coefficients solely as tools for representation decomposition, we interpret them as defining the product law of a feature algebra. Under this perspective, equivariant learning becomes a special instance of feature algebra design within our unified algebraic formalism.

\paragraph{Linear attention and Mamba.}
Softmax attention \citep{vaswani2023attentionneed} has become a fundamental building block of modern deep learning but suffers from quadratic computational complexity. Linear attention \citep{katharopoulos2020transformersrnnsfastautoregressive} addresses this limitation by reformulating attention into a linear-complexity state-update mechanism. State-space models provide another efficient sequence modeling paradigm, and Mamba \citep{gu2023mamba} further introduces input-dependent selection mechanisms that make its computation closely related to linear attention.

Existing work typically explains these architectures in terms of attention mechanisms or state-space dynamics. In contrast, our framework shifts the level of abstraction. We view both linear attention and Mamba as high-level mechanisms emerging from the composition of elementary operators, whose behavior is determined jointly by positional routing and feature algebra. This perspective unifies seemingly different architectures within a common algebraic language.

\section{Algebraic Structure}

\subsection{Tensor product construction }
\label{app: Tensor product}
Let $V$ and $W$ be vector spaces over a field $\mathbb{F}$.
The tensor product $V \otimes W$ is defined as a vector space together with a bilinear map
\[
\otimes : V \times W \to V \otimes W
\]
satisfying the following universal property.

For any vector space $U$ and any bilinear map
\[
B : V \times W \to U,
\]
there exists a unique linear map
\[
\tilde{B} : V \otimes W \to U
\]
such that
 the following diagram commute:
\[
\begin{tikzcd}[row sep=large, column sep=large]
V \times W \arrow[r, "\otimes"] \arrow[dr, "B"'] 
& V \otimes W \arrow[d, "\tilde{B}"] \\
& U
\end{tikzcd}
\]

Explicit construction of the tensor product is the following

Let $F(V \times W)$ denote the free vector space generated by the set $V \times W$.
Let $R \subset F(V \times W)$ be the subspace generated by the relations
\begin{align*}
(v_1 + v_2, w) - (v_1, w) - (v_2, w), \\
(v, w_1 + w_2) - (v, w_1) - (v, w_2), \\
(\alpha v, w) - \alpha (v, w), \\
(v, \alpha w) - \alpha (v, w),
\end{align*}
for all $v, v_1, v_2 \in V$, $w, w_1, w_2 \in W$, and $\alpha \in \mathbb{F}$.

The tensor product is defined as the quotient space
\[
V \otimes W \coloneqq F(V \times W) / R.
\]

The equivalence class of $(v,w)$ in this quotient is denoted by $v \otimes w$.

Let $\{e_i\}_{i=1}^{\dim V}$ be a basis of $V$ and
$\{f_j\}_{j=1}^{\dim W}$ a basis of $W$.
Then the set
\[
\{\, e_i \otimes f_j \mid 1 \le i \le \dim V,\; 1 \le j \le \dim W \,\}
\]
forms a basis of $V \otimes W$.

In particular,
\[
\dim(V \otimes W) = \dim V \cdot \dim W.
\]

Any tensor $T \in V \otimes W$ can be written uniquely as
\[
T = \sum_{i,j} T^{ij} \, e_i \otimes f_j.
\]

Note that $V\otimes W$ is a vector space, while when $V$ and $W$ are algebras, then we can make $V\otimes W$ into algebra by define product as 
\begin{equation*}
    (e_i \otimes f_j)(e_k\otimes f_l)=(e_ie_k)\otimes(f_jf_l)
\end{equation*}
where $e_ie_k$ and $f_jf_l$ are products on $V$ and $W$ respectively. For example if $e_ie_k=\sum_{n}\lambda_{ik}^ne_n$ and $f_jf_l=\sum_m\mu_{jl}^mf_m$. Then 
\begin{equation*}
     (e_i \otimes f_j)(e_k\otimes f_l)=\sum_{n}\sum_m \lambda_{ik}^n\mu_{jl}^m e_n\otimes f_m
\end{equation*}
So the structure constant for algebra $V\otimes W$ is
\begin{equation*}
    \Lambda_{(i,j),(k,l)}^{(n,m)}=\lambda_{ik}^n\mu_{jl}^m.
\end{equation*}

\subsection{Discussion on structure constant}
\label{app: structure constant}
In general, any algebra with basis $\{e_i\}$ is determined by structure constant $\lambda_{ij}^k$
\begin{equation*}
    e_ie_j=\sum_{k}\lambda_{ij}^k e_k
\end{equation*}
Unitality, associativity, and commutativity are encoded in the structure constants.
Existence of unital element $e_{-1}$ that is $e_{-1}e_n=e_n$ corresponds to structure constant $\lambda_{-1,n}^m=\delta_{nm}$.

Associativity $(e_ne_m)e_k=e_n(e_me_k)$ , imposes the following constraint on the structure constants
\begin{equation*}
\sum_m\lambda_{ij}^m\lambda_{mk}^n=\sum_p\lambda_{jk}^p\lambda_{ip}^n \quad \text{(associativity)}.
\end{equation*}
Commutativity $e_ne_m=e_me_n$ imposes the following constraint on the structure constants
\begin{equation*}
    \lambda_{nm}^k=\lambda_{mn}^k \quad \text{(commutativity)}.
\end{equation*}
In our experiment, wo do not impose any constraints.

Matrix Algebra, Clifford Algebra and Algebra with $SO(3)$ representation could also be understand via structure constant but their basis has additional structure and the structure constant are fixed accordingly.

\section{Details}

\subsection{Derivation of convolution}
\label{app: convolution}
The general solution to translation constraint $\lambda(\vec{r}_a, \vec{r}_b; \vec{r}_c )=\lambda(\vec{r}_a, \vec{r}_b+\Delta; \vec{r}_c +\Delta )$ is to let $\lambda$ depends on relative position $\lambda(\vec{r}_a, \vec{r}_b; \vec{r}_c )=\lambda(\vec{r}_a, \vec{r}_b-\vec{r}_c )$. Then we have 
\begin{align*}
    KX&=\sum_c\bigg[\sum_a\sum_b \lambda(\vec{r}_a, \vec{r}_b-\vec{r}_c )K(\vec{r}_a)X(\vec{r}_b)\bigg]f_c\\
    &=\sum_c\bigg[\sum_b \tilde{K}(\vec{r}_b-\vec{r}_c)X(\vec{r}_b)\bigg]f_c
\end{align*}
where we have absorbed $\lambda$ into $\tilde{K}(\vec{r}_b-\vec{r}_c)=\sum_a \lambda(\vec{r}_a, \vec{r}_b-\vec{r}_c )K(\vec{r}_a)$. This is convolution with kernel $\tilde{K}$. So without loss of generality, we can just take $\lambda(\vec{r}_a, \vec{r}_b-\vec{r}_c )=\delta_{\vec{r}_a, \vec{r}_b-\vec{r}_c}$, then the convolution kernel is $K$.

If we want locality, then we can take $K$ satisfy $K(\vec{r}_a)=0$ for $||\vec{r}_a||>R$, where $R$ gives local receptivity field. If there is no locality constraint, this is global convolution and can be efficiently implemented in Fourier space.

\subsection{Clifford convolution}\label{app: Clifford convolution}
For data  represented as
\begin{equation*}
     X=\sum_{a,i}X^i(\vec{r}_a)f_a\otimes e_{i1}\in\mathcal{B}\otimes\mathcal{A}^{\text{Matrix}}\otimes\mathcal{A}^{Cl_2},\, X^i(\vec{r}_a)=X_0^i(\vec{r}_a)+X_1^i(\vec{r}_a)e_1+X_2^i(\vec{r}_a)e_2+X_{12}^i(\vec{r}_a)e_{12}\in \mathcal{A}^{Cl_2}
\end{equation*}
and filter represented as 
\begin{equation*}
    K=\sum_{a,i,j}K^{ij}(\vec{r}_a)f_a\otimes e_{ij}\in\mathcal{B}\otimes\mathcal{A}^{\text{Matrix}}\otimes\mathcal{A}^{Cl_2},\,K^{ij}(\vec{r}_a)=K^{ij}_0(\vec{r}_a)+K^{ij}_1(\vec{r}_a)e_1+K^{ij}_2(\vec{r}_a)e_2+K^{ij}_{12}(\vec{r}_a)e_{12}
\end{equation*}

we have Clifford convolution operator
\begin{equation*}
\mathcal{O}^{\text{Conv},\mathcal{A}^{\text{Matrix}}\otimes\mathcal{A}^{Cl_2}}(K,X)=\sum_{c,i}\bigg[\sum_b\sum_jK^{ij}(\vec{r}_b-\vec{r}_c)X^j(\vec{r}_b)\bigg]f_c\otimes e_{i1}
\end{equation*}
where
\begin{align*}
    &K^{ij}(\vec{r}_b-\vec{r}_c)X^j(\vec{r}_b)\\
    &\quad=K_0^{ij}(\vec{r}_b-\vec{r}_c)X_0^j(\vec{r}_b) +K_1^{ij}(\vec{r}_b-\vec{r}_c)X_1^j(\vec{r}_b)+K_2^{ij}(\vec{r}_b-\vec{r}_c)X_2^j(\vec{r}_b)-K_{12}^{ij}(\vec{r}_b-\vec{r}_c)X_{12}^j(\vec{r}_b)\\
&\qquad+\bigg[K_0^{ij}(\vec{r}_b-\vec{r}_c)X_1^j(\vec{r}_b) +K_1^{ij}(\vec{r}_b-\vec{r}_c)X_0^j(\vec{r}_b)-K_2^{ij}(\vec{r}_b-\vec{r}_c)X_{12}^j(\vec{r}_b)+K_{12}^{ij}(\vec{r}_b-\vec{r}_c)X_{2}^j(\vec{r}_b)\bigg]e_1\\
&\qquad+\bigg[K_0^{ij}(\vec{r}_b-\vec{r}_c)X_2^j(\vec{r}_b) +K_1^{ij}(\vec{r}_b-\vec{r}_c)X_{12}^j(\vec{r}_b)+K_2^{ij}(\vec{r}_b-\vec{r}_c)X_{0}^j(\vec{r}_b)-K_{12}^{ij}(\vec{r}_b-\vec{r}_c)X_{1}^j(\vec{r}_b)\bigg]e_2\\
&\qquad+\bigg[K_0^{ij}(\vec{r}_b-\vec{r}_c)X_{12}^j(\vec{r}_b) +K_1^{ij}(\vec{r}_b-\vec{r}_c)X_{2}^j(\vec{r}_b)-K_2^{ij}(\vec{r}_b-\vec{r}_c)X_{1}^j(\vec{r}_b)+K_{0}^{ij}(\vec{r}_b-\vec{r}_c)X_{12}^j(\vec{r}_b)\bigg]e_1e_2
\end{align*}

\subsection{Equivariant Networks}

\paragraph{Geometry transformation}
\label{app: geometry transformation}
Let $M=\mathbb{R}^3$ and consider a trivial vector bundle
$E=M\times V$, where $V$ carries a (finite–dimensional) representation of
$SO(3)$, denoted $\rho(g):V\to V$.  In trivial bundle case, sections of $E$ is $\Gamma(E)=\{s:M\to V\}$. Bundle  homomorphism $T_g(p,v_p)=(gp,\rho(g)v_p)$   induce action  of $SO(3)$ on sections by
\begin{equation*}
    (T_{R}s)(\vec r)
      := \rho(R)\,s(R^{-1}\vec r),
      \qquad R\in SO(3).
    \label{eq:section-action}
\end{equation*}
An operator $\mathcal{L}:\Gamma(E)\to\Gamma(E)$ is $SO(3)$–equivariant  if
\begin{equation*}
    T_{R}\mathcal{L}(s) = \mathcal{L}(T_{R}s)
    \qquad\forall\,s\in\Gamma(E),\; R\in SO(3).
    \label{eq:equiv-def}
\end{equation*}

We decompose $V$ as a direct sum of irreducible $SO(3)$ representations,
$
   V = \bigoplus_{l\in\mathbb{N}} V_l,
   \;
   V_l = \mathrm{span}\{e^{l}_{m}\}_{m=-l,\dots,l},
$
so that for each $R\in SO(3)$
\[
   \rho(R)e^{l}_{m}
    = \sum_{m'=-l}^{l} D^{(l)}_{m'm}(R)\,e^{l}_{m'},
\]
where $D^{(l)}(R)$ is the Wigner–$D$ matrix.  Under this decomposition a field
$s$ can be written as
$
   s(\vec r)
   = \big( s^{l}_{m}(\vec r) \big)_{l,m},
   \;
   s^{l}_{m} : \mathbb{R}^{3}\to\mathbb{R}.
$

Follow the convention of tensor field network we consider action of rotation group as
\begin{equation}
    (T_{R}s)^{\,l}_{m}(\vec r)
      = \sum_{m'=-l}^{l} D^{(l)}_{m' m}(R)\,
        s^{l}_{m'}(R^{-1}\vec r).
    \label{eq:section-action-components}
\end{equation}

\paragraph{Equivariance condition on $\mathcal{O}^{\text{conv},\mathcal{A}^{\text{SO}(3)}}$}
\label{app: SO(3) equivariance}

The derivation here follows \citep{dong2026productinteractionalgebraicformalism}

For a discretized field $X=\{X_m^l\}$ represented in algebra $\mathcal{B}\otimes\mathcal{A}^{SO(3)}$ as $X=\sum_a\sum_{l,m}X_m^l(\vec{r}_a)f_a\otimes e_m^l$, 
We define its transformation follow geometry definition (\eqref{eq:section-action-components})  and transfer it to representation of feature algebra.
\begin{align}
    \mathcal{T}_R(X)&=\sum_{a}\sum_l\sum_{m=-l}^l (T_RX)_m^l(\vec{r_a})f_a\otimes  e_m^l \notag\\
    &=\sum_{a}\sum_l\sum_{m=-l}^l\sum_{m'=-l}^lD(R)_{m'm}^{(l)}X^l_{m'}(R^{-1}\vec{r_a}) f_a\otimes e_m^l \notag\\
    &=\sum_{a}\sum_l\sum_{m'=-l}^lX^l_{m'}(R^{-1}\vec{r_a}) f_a\otimes (\rho(R^{-1})^*e_{m'}^l) \label{eq:lift transformation}
\end{align}
Where $\rho(R^{-1})^*e_{m'}^l=\sum_{m=-l}^lD(R^{-1})^{(l)*}_{mm'}e_m^l=\sum_{m=-l}^lD(R)_{m'm}^{(l)}e_m^l$ and we use the unitary condition of $D(R)$.

The equivariance condition is
\begin{equation*}
\mathcal{T}_R\big(\mathcal{O}^{\text{Conv},\mathcal{A}^{SO(3)}}(K,X)\big)=\mathcal{O}^{\text{Conv},\mathcal{A}^{SO(3)}}\big(K,\mathcal{T}_R(X)\big)\qquad R\in SO(3)
\end{equation*}

Note that
\begin{align}
    \sum_{m=-l}^l K_m^l e_m^l&=\sum_mK_m^l[\rho(R^{-1})^*\rho(R)^*e_m^l]\notag\\
&=\sum_{m}K_m^l[\sum_{m'}D(R)^{(l)*}_{m'm}\sum_nD(R^{-1})^{(l)*}_{nm'}e_n^l]\notag\\
&=\sum_{m'}(\sum_mD(R)^{(l)*}_{m'm}K_m^l)(\sum_nD(R^{-1})^{(l)*}_{nm'}e_n^l)\notag\\
&=\sum_{m'}(\sum_mD(R^{-1})^{(l)}_{mm'}K_m^l)(\rho(R^{-1})^*e_{m'}^l) \label{eq: expression for K}
\end{align}

Then use \eqref{eq:lift transformation} and \eqref{eq: expression for K} we have
\begin{align*}
&K\mathcal{T}_R(X)\\
&=\sum_{a,b}\sum_{l_1,m_1}\sum_{l_2,m_2}(\sum_{m=-l_1}^{l_1}D(R^{-1})_{mm_1}^{l_1}K_{m}^{l_1}(\vec{r_a}-\vec{r_b}))X_{m_2}^{l_2}(R^{-1}\vec{r_b})f_a\otimes  \rho(R^{-1})^*e_{m_1}^{l_1}\rho(R^{-1})^*e_{m_2}^{l_2}\\
&=\sum_{a,b}\sum_{l_1,m_1}\sum_{l_2,m_2}(\sum_{m=-l_1}^{l_1}D(R^{-1})_{mm_1}^{l_1}K_{m}^{l_1}(\vec{r_a}-\vec{r_b}))X_{m_2}^{l_2}(R^{-1}\vec{r_b})f_a\otimes  \rho(R^{-1})^*(e_{m_1}^{l_1}e_{m_2}^{l_2})
\end{align*}
where we have used the consist condition $(\rho(R)e^{l_1}_{m_1})(\rho(R)e^{l_2}_{m_2})
   = \rho(R)\big(e^{l_1}_{m_1} e^{l_2}_{m_2}\big)$.
If we have
\begin{equation*}
    \sum_{m=-l_1}^{l_1}D(R^{-1})_{mm_1}^{l_1}K_{m}^{l_1}(\vec{r_a}-\vec{r_b})=K_{m_1}^{l_1}(R^{-1}\vec{r_a}-R^{-1}\vec{r_b})
\end{equation*}

Then
\begin{align*}
&K\mathcal{T}_R(X)\\
&=\sum_{a,b}\sum_{l_1,m_1}\sum_{l_2,m_2}K_{m_1}^{l_1}(R^{-1}\vec{r_a}-R^{-1}\vec{r_b})X_{m_2}^{l_2}(R^{-1}\vec{r_b})f_a\otimes  \rho(R^{-1})^*(e_{m_1}^{l_1}e_{m_2}^{l_2})\\
&=\mathcal{T}_R(KX)
\end{align*}
which gives equivariance.

So equivariance condition forces the kernel to satisfy
\begin{equation}
   \sum_{m=-l}^{l}
     D^{(l)}_{m m_1}(R^{-1})
     K^{\,l}_{m}(\vec r)
   =
   K^{\,l}_{m_1}(R^{-1}\vec r),
   \qquad\forall R\in SO(3).
   \label{eq:K-equiv}
\end{equation}
Solutions of \eqref{eq:K-equiv} are exactly of the form
\[
 K^{\,l}_{m}(\vec r)
   = R^{l}(\|\vec r\|)\,Y^{l}_{m}(\hat r),
\]
with $Y^{l}_{m}$ spherical harmonics and $R^{l}$ an arbitrary radial profile.
Substituting into basic product operator
\begin{equation*}
\mathcal{O}^{\text{Conv},\mathcal{A}^{SO(3)}}(K,X)=\sum_{c;\,l,m}\bigg[\sum_b\sum_{l_2,m_2}\bigg(\sum_{l_1,m_1}C_{l_1,m_1;l_2,m_2}^{l,m}K_{m_1}^{l_1}(\vec{r}_b-\vec{r}_c)\bigg)X_{m_2}^{l_2}(\vec{r}_b)\bigg]f_c\otimes e_m^l
\end{equation*}
recovers the standard Tensor Field network.

\subsection{Compositional construction of Mamba}\label{app: Mamba}
For data represented as 
$$X=\sum_{k;\,\alpha} X_\alpha(k) f_k\otimes e_{ \alpha 1}\in \mathcal{B}\otimes\mathcal{A}^{\text{Matrix}},$$
 we get $B$, $C$ and $\Delta$ through linear operator $\text{id}\otimes \mathcal{W}^B$, $\text{id}\otimes \mathcal{W}^C$ and $\text{id}\otimes \mathcal{W}^\Delta$ and nonlinear activation $F$ which is softplus.
\begin{align*}
    B&= \text{id}\otimes \mathcal{W}^B(X)=\sum_{k;\,i}\big(\sum_\beta W^B_{i\beta}X_\beta(k)\big)f_k\otimes e_{1i}\\
    C&= \text{id}\otimes \mathcal{W}^C(X)=\sum_{k;\,i}\big(\sum_\beta W^C_{i\beta}X_\beta(k)\big)f_k\otimes e_{i1}\\
    \Delta &= F\big(\text{id}\otimes \mathcal{W}^\Delta(X)\big)=\sum_{k;\,\alpha,i}\ F\big(\sum_\beta W^\Delta_{\alpha\beta}X_\beta(k)\big)f_k\otimes e_{\alpha i}
\end{align*}
where $\mathcal{W}^Be_{\beta 1}=\sum_iW^B_{i\beta}e_{1i}$, $\mathcal{W}^Ce_{\beta 1}=\sum_iW^C_{i\beta}e_{i1}$ and $\mathcal{W}^\Delta e_{\beta1}=\sum_i\sum_\alpha W^\Delta_{\alpha\beta}e_{\alpha i}$.

We first get $I$ via local-local interaction operator.
\begin{align*}
    I_0&=\mathcal{O}_{\, \mathbb{R}^d\times\mathbb{R}^d\to\mathbb{R}^{d\times d}}^{L,\mathcal{A}^{\text{Matrix}}}(X,B)=\sum_{k;\,\alpha, i}\bigg[B_i(k)X_\alpha(k)\bigg]f_k\otimes e_{\alpha i},\\
    I&=\mathcal{O}_{\, \mathbb{R}^{d\times d}\times\mathbb{R}^{d\times d}\to\mathbb{R}^{d\times d}}^{L,\mathcal{A}^{\text{Hadamard}}}(\Delta,I_0)=\sum_{k;\,\alpha, i}\bigg[\Delta_{\alpha i}(k)B_i(k)X_\alpha(k)\bigg]f_k\otimes e_{\alpha i}\\
\end{align*}
And $A$ is 
\begin{equation*}
A=\mathcal{O}^{\mathcal{A}^\text{Hadamard}}_{\mathbb{R}^{d\times d}}(\Delta,\lambda)=\sum_{k;\,\alpha, i}\bigg[\lambda_{\alpha i}\Delta_{\alpha i}(k)\bigg]f_k\otimes e_{\alpha i}
\end{equation*}
Now we construct hidden state $H$ via causal local-global interaction operator
\begin{align*}
H^{\text{local}}&=\mathcal{O}^{G,\mathcal{A}^\text{Hadamard}}_{F,\mathbb{R}^{d\times d}\times\mathbb{R}^{d\times d}\to\mathbb{R}^{d\times d}}(I,A)=\sum_{k;\,\alpha, i}\bigg[F\bigg(\sum_{l=1}^kA_{\alpha i}(l)\bigg)I_{\alpha i}(k)\bigg]f_k\otimes e_{\alpha i},\quad F(x)=\exp(-x)\\
H^{\text{global}}&=\mathcal{O}^{G,\mathcal{A}^\text{Hadamard}}_{\mathbb{R}^{d\times d}}(1,H^{\text{local}})=\sum_{t;\,\alpha, i}\bigg[\sum_{k=1}^tH^{\text{local}}_{\alpha i}(k)\bigg]f_t\otimes e_{\alpha i}\\
H&=\mathcal{O}^{G,\mathcal{A}^\text{Hadamard}}_{F,\mathbb{R}^{d\times d}\times\mathbb{R}^{d\times d}\to\mathbb{R}^{d\times d}}(H^{\text{global}},A)=\sum_{t;\,\alpha, i}\bigg[F\bigg(\sum_{l=1}^tA_{\alpha i}(l)\bigg)H^{\text{global}}_{\alpha i}(t)\bigg]f_t\otimes e_{\alpha i},\quad F(x)=\exp(x)
\end{align*}
Note that the three local-global operator corresponds exactly to the three integration to solve the ODE. The final output is through local-local operator
\begin{equation*}
    Y = \mathcal{O}^{L,\mathcal{A}^{\text{Matrix}}}_{\mathbb{R}^{d\times d}\times\mathbb{R}^{d}\to\mathbb{R}^{d}}(H,C)=\sum_{t,\alpha}\bigg[\sum_iH_{\alpha i}(t)C_i(t)\bigg]f_t\otimes e_{\alpha 1}
\end{equation*}

\section{Experimental details}

\subsection{Experimental details of lambda-Clifford interaction }
\label{app: Experiment lambda-Clifford}
The linear attention baseline is obtained by replacing the standard $3 \times 3$ convolution in the secondary layer of each standard BasicBlock of ResNet-20 by a standard Linear Attention Layer which is global attention due to linear complexcity. For an input feature map $X$, the query ($Q$), key ($K$), and value ($V$) projections are generated via $1 \times 1$ convolutions. To compensate for the loss of spatial locality inherent in global attention, an Implicit Position Encoding (IPE) module—implemented via a $3 \times 3$ depthwise separable convolution with padding—is seamlessly integrated into the value projection path.

For $\lambda$-Clifford model, we fix all the other components and only replace linear attention block by $\lambda$- Clifford interaction.  

We evaluated  models on the benchmark CIFAR-10 dataset. Input images were dynamically resized to a resolution of $72 \times 72$ pixels. The original training set was partitioned into a 90/10 split for training and validation, respectively.

All the networks were optimized using the AdamW optimizer with an initial learning rate of $1 \times 10^{-3}$ and a weight decay coefficient of $0.01$. The learning rate was decayed following a Cosine Annealing schedule down to a minimum value of $1 \times 10^{-6}$ over a maximum of 200 epochs. Training was conducted with a batch size of 128 utilizing cross-entropy loss. All training and inference experiments were conducted on an NVIDIA RTX 4090 GPU with 48 GB of VRAM and a 20‑core CPU.
We run 5 random seeds and for each seed, 
model checkpoints were monitored based on validation accuracy, and the optimal weights were selected to report the final inference performance on the independent test dataset. Then we obtain mean and standard deviations from results of 5 random seeds.

\subsection{Experimental details of vision Mamba and $\lambda$ SSM}
\label{app: experiment lambda SSM}
For Vision Mamba baseline,
we adopt a multi-head bidirectional Vision Mamba classifier for CIFAR-10. Input images (32×32) are split into non-overlapping patches of size 4×4 via a convolutional patch embedding, yielding a sequence of 64 tokens with channel dimension $d_{\text{model}}=128$.
 A learnable positional encoding is added to the patch tokens before they are passed through a stack of 8 VMamba blocks. Each block consists of a residual connection around a pre-norm RMSNorm and a VMamba mixer. The mixer expands the input to an inner dimension $2\cdot d_{\text{model}}=256$
 via a linear projection, splits it into feature and gating branches, and applies a causal 1D convolution (kernel size 4). The core SSM dynamical layer is extended with $n_{\text{heads}}=4$ heads
( $d_\text{head}=64$, state dimension $d_{\text{hidden}}=16$, where the input-dependent parameters $\Delta$, $B$ and $C$
are produced by a linear layer and reshaped into multi-head tensors.  To capture bidirectional context, a mirrored backward branch with independent parameters processes the reversed sequence, and the two outputs are averaged. After the VMamba stack, we apply global average pooling, an RMSNorm, and a linear classification head for 10 classes.

For $\lambda$-SSM model, we fix all the other part of VMamba architecture and only
 replace the mamba dynamical layer with our $\lambda$-SSM dynamical layer.
We follow exactly the same hyperparameter as Vision Mamba. $B$ and $C$ are project to each head with dimension $d_{\text{hidden}}=16$ and $X$ is reshaped into each head with $d_\text{head}=64$, then $X$ and $B$ are multiplied to form hidden state $H$ with dimension $d_{\text{hidden}}=16$ in each head via shared structure constant $\lambda$. The final output is get from product between $H$ and $C$ in each head via another shared structure constant $\lambda$. 
Note that the hidden state dimension in the $\lambda$-SSM is only 16, whereas that in Mamba is $64\times 16$
(i.e., 1024). Despite this substantial reduction in dimensionality, the performance degradation remains modest.

Training is performed on CIFAR-10 with standard data augmentation. Inputs are normalized with dataset statistics. We use the AdamW optimizer with a learning rate of $1\times 10^{-3}$ and weight decay 0.05, together with a cosine annealing rschedule decaying to $1\times 10^{-6}$ over 300 epochs. The batch size is 128, and 10 percentage of the training set is held out for validation. The validation dataset is not used for training. We run 5 random seeds and select checkpoints with best validation accuracy for each seed and record corresponding test accuracy. Then we calculate mean and stand standard deviations. All training and inference experiments were conducted on an NVIDIA RTX 4090 GPU with 48 GB of VRAM and a 20‑core CPU. Training and inference speed measurements are reported based on runs performed on this instance under identical training configurations.

\subsection{Experimental details of  Clebsch–Gordan Interaction }
\label{app:experiment_Clebsch–Gordan}
Expression for Clebsch–Gordan Interaction  is
\begin{align}\label{eq: Clebsch–Gordan Interaction}
    H&=\mathcal{O}^{L,\mathcal{A}^{SO(3)}}(K,V)=\sum_{\vec{r}_b}\sum_{l_1,m_1}\bigg[\sum_{l_3,m_3}\sum_{l_4,m_4}C_{l_4,m_4;l_3,m_3}^{l_1,m_1}K_{m_4}^{l_4}(\vec{r}_b)V_{m_3}^{l_3}(\vec{r}_b)\bigg]f_b\otimes e_{m_1}^{l_1}\\
    O&=\mathcal{O}^{G,\mathcal{A}^{SO(3)}}(Q,H)=\sum_{\vec{r}_a}\sum_{l,m}\bigg[\sum_{l_1,m_1}\sum_{l_2,m_2}C_{l_1,m_2;l_2,m_2}^{l,m}Q_{m_2}^{l_2}(\vec{r}_a)\bigg(\sum_{\vec{r}_b}H_{m_1}^{l_1}(\vec{r}_b)\bigg)\bigg]f_a\otimes e_m^l
\end{align}

\begin{figure}[htbp]
  \centering
  \begin{minipage}{0.42\textwidth}
    \centering
    \includegraphics[width=\linewidth]{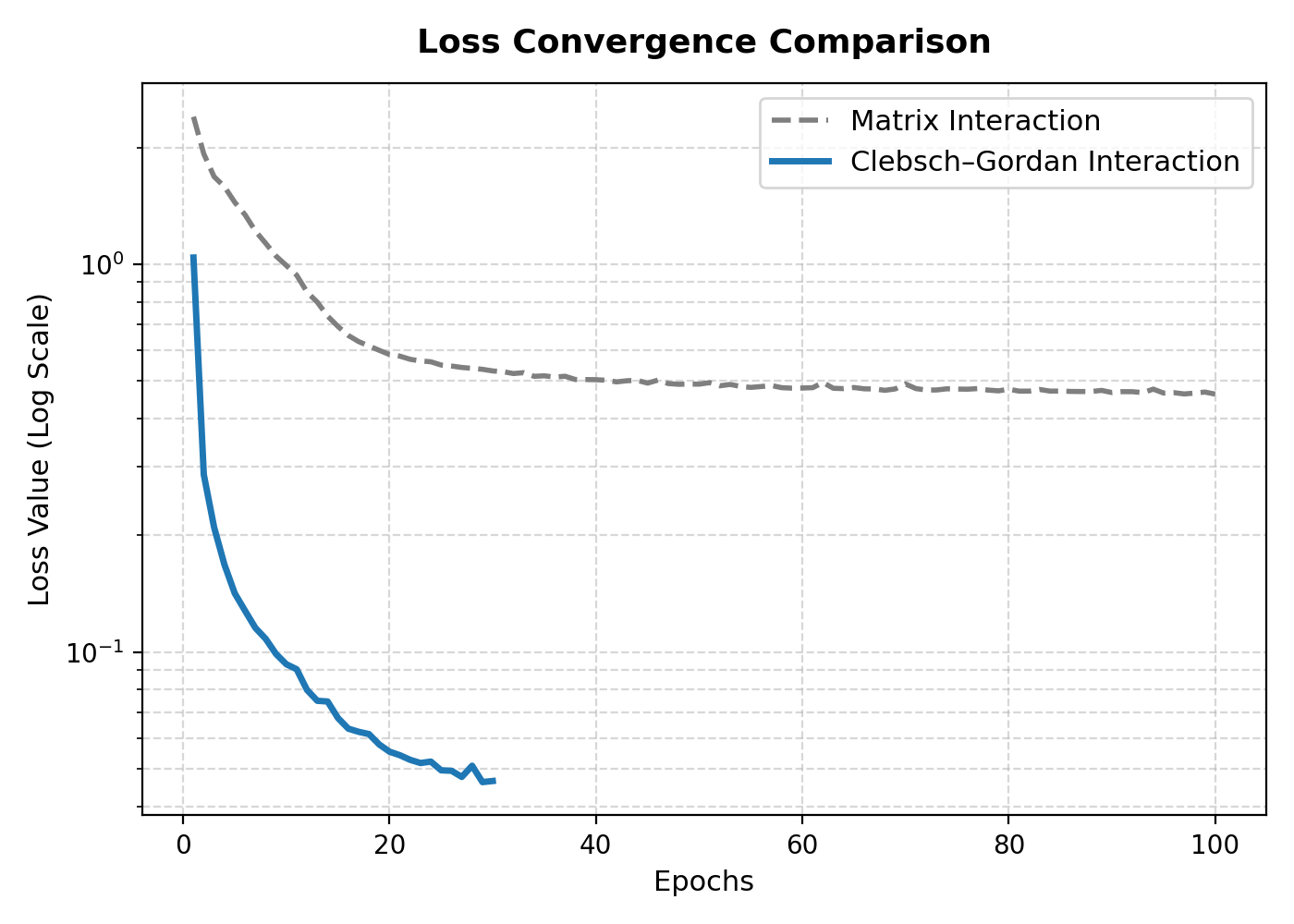}
    \caption{Loss}
    \label{fig:so3_loss}
  \end{minipage}\hfill
  \begin{minipage}{0.42\textwidth}
    \centering
    \includegraphics[width=\linewidth]{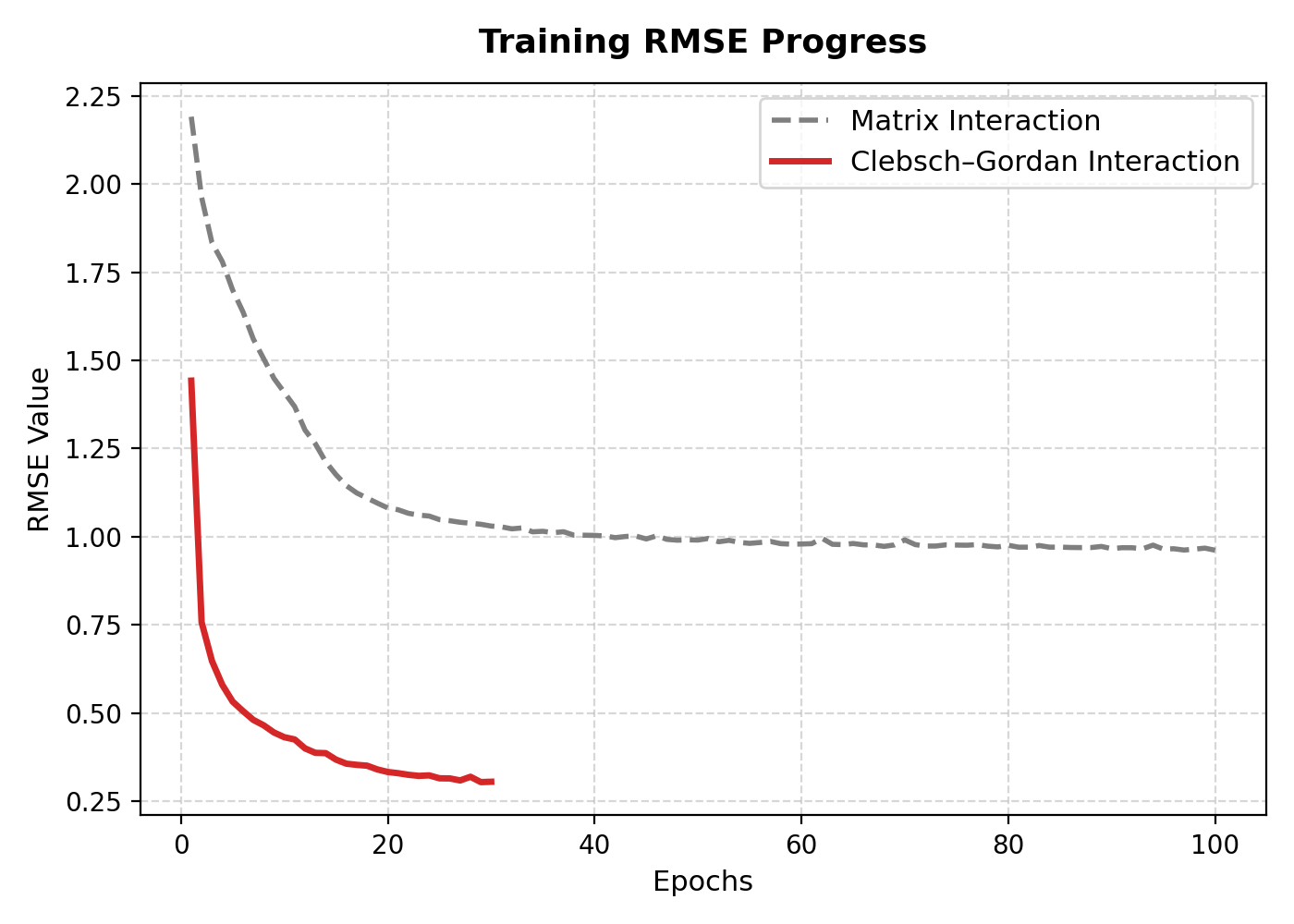}
    \caption{RMSE}
    \label{fig:so3_RMSE}
  \end{minipage}
\end{figure}

The experimental setup follows missing point task in \citep{thomas2018tensor}. The initial sparse categorical node representation  goes through a self-interaction layer acting as a dense embedding matrix to formulate feature with channel capacity $16$. The architecture stacks 3 identical, progressive interaction layer with channel capacity $16-16-16$. For each interaction layer, we get $Q$ from equivariant TFN convolution and get $K$ and $V$ through self-interaction layer. Then for Clebsch–Gordan Interaction, QKV are mixed in their representation dimension as \eqref{eq: Clebsch–Gordan Interaction}. While for Matrix interaction, we flatten channel dimension and representation dimension and mix QKV via usual linear attention. So all the other components are same for Clebsch–Gordan Interaction architecture and Matrix interaction architecture and
the only difference lies in how QKV are mixed, for Clebsch–Gordan Interaction, the mixing is in algebra $\mathcal{A}^{SO(3)}$ which preserve symmetry while for Matrix interaction the mixing is in algebra $\mathcal{A}^{\text{Matrix}}$ which destroy symmetry. The training curve is shown in Figures~\ref{fig:so3_loss}and \ref{fig:so3_RMSE} and the final test results are shown in Tables\ref{tab:Accuracy by Distance}and \ref{tab:Accuracy of Atom Type}

\begin{table}[htbp]
\centering
\caption{Accuracy by Distance }
\label{tab:Accuracy by Distance}
\small
\begin{tabular}{lcc}
\toprule
\textbf{Model} & Clebsch–Gordan Interaction(epoch 10) & Matrix Interaction(epoch 100)  \\
\midrule
5-18 Atoms  & 0.9396 &  0.3493  \\
19 Atoms & 0.9567 & 0.3146  \\
23 Atoms & 0.9587  & 0.2836   \\
29 Atoms & 0.9532      & 0.2554\\
\bottomrule
\end{tabular}
\end{table}

\begin{table}[htbp]
\centering
\caption{Accuracy of Atom Type}
\label{tab:Accuracy of Atom Type}
\small
\begin{tabular}{lcc}
\toprule
\textbf{Model} & Clebsch–Gordan Interaction(epoch 10) & Matrix Interaction(epoch 100)  \\
\midrule
5-18 Atoms  & 0.9870 &  0.8897  \\
19 Atoms & 0.9773 & 0.8931  \\
23 Atoms & 0.9791  & 0.9252   \\
29 Atoms & 0.9852      & 0.9601\\
\bottomrule
\end{tabular}
\end{table}

Figures~\ref{fig:so3_loss} and~\ref{fig:so3_RMSE} show a clear difference in optimization behavior between the two models. The model based on Clebsch--Gordan interactions converges substantially faster and reaches a significantly lower loss and RMSE than the matrix-interaction baseline. As a consequence, the Clebsch--Gordan model already achieves strong performance after only 10 training epochs, whereas the matrix-interaction model fails at this task.

\begin{figure}[htbp]
  \centering
  \begin{minipage}{0.48\textwidth}
    \centering
    \includegraphics[width=\linewidth]{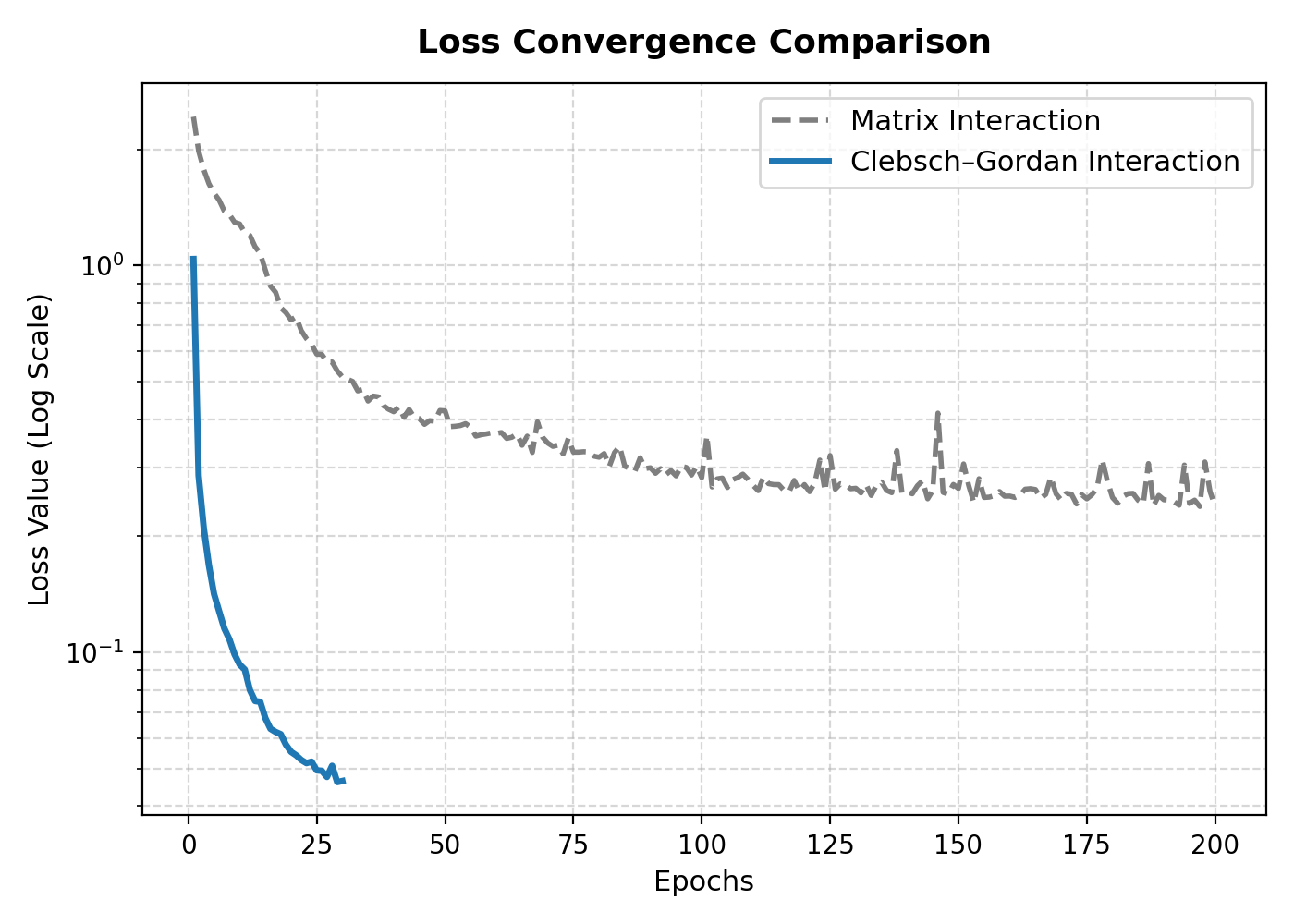}
    \caption{Loss}
    \label{fig:so3_loss2}
  \end{minipage}\hfill
  \begin{minipage}{0.48\textwidth}
    \centering
    \includegraphics[width=\linewidth]{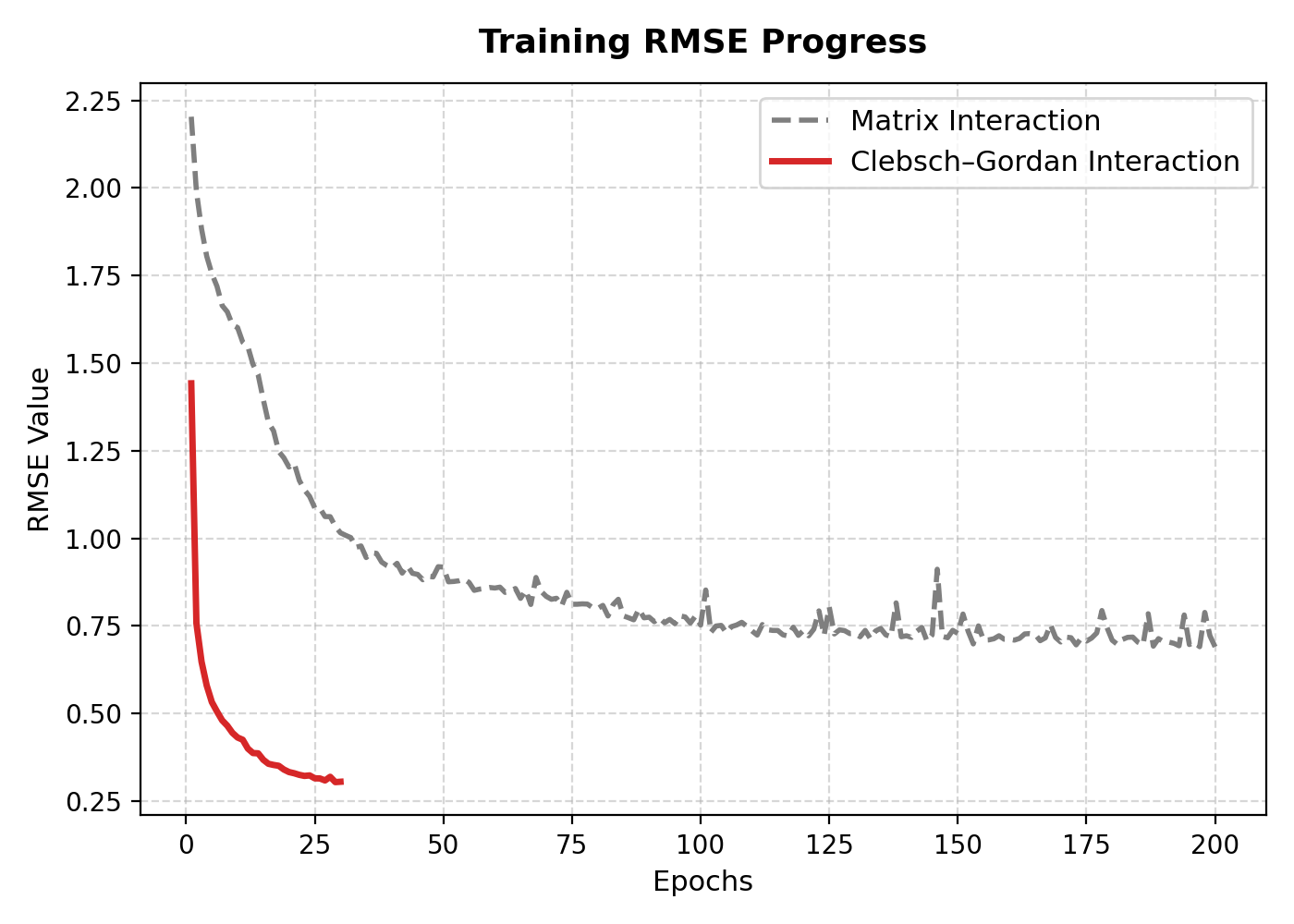}
    \caption{RMSE}
    \label{fig:so3_RMSE2}
  \end{minipage}
\end{figure}

In the above experimental setup, we isolate influence of algebraic mixing rule and keep all the other components same. We also consider a standard linear attention architecture for this task. In this setup, categorical one-hot profile and its absolute Cartesian vector are explicitly smashed into a unified, flat continuous vector, then the vector is embed to feature vector with dimension $16$ via linear embed layer. Then the feature vector goes through
3 standard transformer block with linear attention and the final output are project to make prediction. Training cure for this standard linear attention architecture is shown in Figure~\ref{fig:so3_loss2} and ~\ref{fig:so3_RMSE2} and the test result are shown in Table\ref{tab:Accuracy by Distance2}

\begin{table}[htbp]
\centering
\caption{Accuracy by Distance  }
\label{tab:Accuracy by Distance2}
\small
\begin{tabular}{lcc}
\toprule
\textbf{Model} & Clebsch–Gordan Interaction(epoch 10) & Matrix Interaction(epoch 200)  \\
\midrule
5-18 Atoms  & 0.9396 & 0.8373  \\
19 Atoms & 0.9567 & 0.8155  \\
23 Atoms & 0.9587  & 0.7910   \\
29 Atoms & 0.9532      & 0.8287\\
\bottomrule
\end{tabular}
\end{table}
The standard linear attention block is found to be less accurate and more slowly converging than the Clebsch–Gordan Interaction. This further shows the importance of encoding symmetry in algebraic interaction.

\end{document}